\documentclass{article}

    \PassOptionsToPackage{numbers, compress}{natbib}

\usepackage[final]{neurips_2024}

\usepackage{multirow} 
\usepackage{graphicx}
\usepackage{algorithm, algorithmic}
\usepackage{colortbl}
\usepackage{amsmath}
\usepackage{amssymb}
\usepackage{wrapfig} 
\usepackage{enumerate}

\usepackage[utf8]{inputenc} 
\usepackage[T1]{fontenc}    
\usepackage{hyperref}       
\usepackage{url}            
\usepackage{booktabs}       
\usepackage{amsfonts}       
\usepackage{nicefrac}       
\usepackage{microtype}      
\usepackage{xcolor}         

\title{RG-SAN: Rule-Guided Spatial Awareness Network for End-to-End 3D Referring Expression Segmentation}

\author{%
  Changli Wu$^{1,2}$\thanks{Equal Contribution.}\;\,, Qi Chen$^{1*}$, Jiayi Ji$^{1,4*}$, Haowei Wang$^{3}$, Yiwei Ma$^{1}$, \\
  \textbf{You Huang$^{1}$, Gen Luo$^{1}$, Hao Fei$^{4}$, Xiaoshuai Sun$^{1}$, Rongrong Ji$^{1}$}\thanks{Corresponding Author.}\\
  $^{1}$ Key Laboratory of Multimedia
  Trusted Perception and Efficient Computing, \\
  Ministry of Education of China,
  Xiamen University,
  361005, P.R. China \\
  $^{2}$ Shanghai Innovation Institute, Shanghai, P.R. China \\
  $^{3}$ Youtu Lab, Tencent, Shanghai, P.R. China \\
  $^{4}$ National University of Singapore
}

\begin{document}

\maketitle

\begin{abstract}
3D Referring Expression Segmentation (3D-RES) aims to segment 3D objects by correlating referring expressions with point clouds. However, traditional approaches frequently encounter issues like over-segmentation or mis-segmentation, due to insufficient emphasis on spatial information of instances. In this paper, we introduce a Rule-Guided Spatial Awareness Network  (RG-SAN) by utilizing solely the spatial information of the target instance for supervision. This approach enables the network to accurately depict the spatial relationships among all entities described in the text, thus enhancing the reasoning capabilities. The RG-SAN consists of the Text-driven Localization Module (TLM) and the Rule-guided Weak Supervision (RWS) strategy. The TLM initially locates all mentioned instances and iteratively refines their positional information. The RWS strategy, acknowledging that only target objects have supervised positional information, employs dependency tree rules to precisely guide the core instance's positioning. Extensive testing on the ScanRefer benchmark has shown that RG-SAN not only establishes new performance benchmarks, with an mIoU increase of 5.1 points, but also exhibits significant improvements in robustness when processing descriptions with spatial ambiguity. All codes are available at~\url{https://github.com/sosppxo/RG-SAN}.
\end{abstract}

\section{Introduction}
\label{sec:intro}
3D Referring Expression Segmentation (3D-RES) is an emerging field that segments 3D objects in point cloud scenes based on given referring expressions~\cite{Huang_Lee_Chen_Liu_2021}. Gaining significant attention for its applications in autonomous robotics, human-machine interaction, and self-driving systems, 3D-RES demands a deeper understanding than 3D Referring Expression Comprehension (3D-REC)~\cite{Chen_Chang_Nießner_2020, Yuan_Yan_Liao_Zhang_Wang_Li_Cui_2021, Achlioptas_Abdelreheem_Xia_Elhoseiny_Guibas_2020, Zhao_Cai_Sheng_Xu_2021,Wu_Cheng_Zhang_Cheng_Zhang_2022}, which focuses only on locating the referring objects via bounding boxes. 3D-RES, on the other hand, requires identifying instances and providing precise 3D masks.

\begin{wrapfigure}{r}{0.5\textwidth}
\centering 
  \includegraphics[width=1.\linewidth]{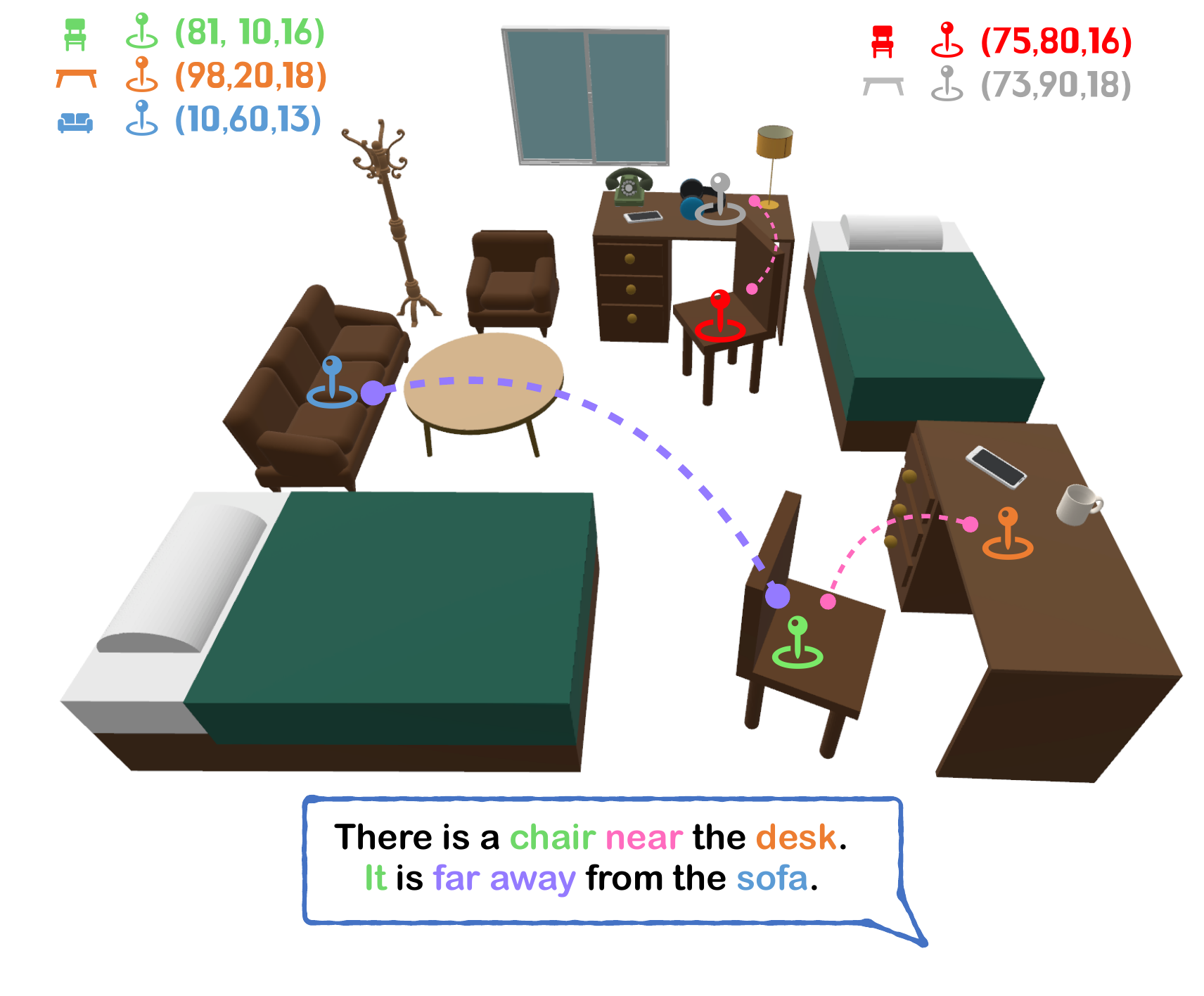}
  \caption{Illustration with a target object and multiple auxiliary objects, associated with a referring expression. The target marked in green represents the main referred instance, while targets in other colors indicate other mentioned entities. This visual highlights the challenge of effectively completing semantic reasoning in the absence of spatial inference.
  }
  \label{fig:1}
\end{wrapfigure}

Early 3D-RES approaches~\cite{Huang_Lee_Chen_Liu_2021, Yuan_Yan_Liao_Zhang_Wang_Li_Cui_2021} adopted a two-stage paradigm, starting with an independent text-agnostic segmentation model for generating instance proposals, followed by linking these proposals with textual descriptions. This paradigm, separating segmentation and matching, proved suboptimal in performance and efficiency. Recent explorations have shifted towards an end-to-end paradigm. For instance, 3D-STMN~\cite{wu20233d} achieved efficient segmentation by directly matching superpoints with text, while 3DRefTR~\cite{lin2023unified} integrated 3D-RES and 3D-REC into a unified framework using a multi-task approach, boosting inference in both tasks. Despite these advancements, limitations persist, primarily due to over-reliance on textual reasoning and insufficient modeling of spatial relationships between instances. 
For example, as shown in Fig.~\ref{fig:1}, without spatial modeling, it's challenging to understand and correctly segment the intended chair in scenarios involving complex spatial terms like ``far away''.

To tackle this issue, the core is to assist textual reasoning by modeling the spatial relationships of core instances. 
By effectively identifying these spatial relationships within expressions, a substantial improvement can be achieved in comprehending spatial arrangements.
Nevertheless, this endeavor is not without its challenges. While accurate positional information is crucial for ensuring precise modeling of spatial relationships, accurately regressing instance positions from textual information is far from a simple task. Furthermore, our available positional information is limited to the target instance, leaving us without supervisory signals for other instances referenced in the expression.

To overcome these challenges, we propose the novel Rule-Guided Spatial Awareness Network (RG-SAN), utilizing the spatial information of the target instance for supervision. This enables the network to accurately depict spatial relationships among all text-described entities, thereby significantly enhancing the model's inference and pointing capabilities. RG-SAN consists of two main components: the Text-driven Localization Module (TLM) and the Rule-guided Weak Supervision (RWS) strategy. TLM initially locates all mentioned instances and iteratively refines their positions, ensuring continuous improvement in location accuracy. RWS, leveraging dependency tree rules, precisely guides the positioning of core instances. This focused supervision significantly improves the handling of spatial ambiguities in referring expressions. Extensive testing on the ScanRefer benchmark shows that RG-SAN not only sets new performance standards, with a mIoU increase of 5.1 points, but also greatly enhances robustness in processing spatially ambiguous descriptions.

To sum up, our main contributions are as follows:
\begin{itemize}
    \item We introduce RG-SAN, a novel approach for modeling spatial relationships among all entities in expressions, which enhances the model's referring ability in 3D-RES.
    \item We propose the TLM for precise localization of all instances mentioned in expressions, and RWS, utilizing only the target instance's location for supervising the spatial positioning of all instances.
    \item Extensive experiments on the ScanRefer benchmark demonstrate the effectiveness of the proposed RG-SAN, showing significant improvements in performance and robustness in 3D-RES tasks.
\end{itemize}

\section{Related Work}
\label{sec:related_work}

\subsection{3D Referring Expression Comprehension and Referring Expression Segmentation}
Referring Expression Comprehension (REC) is proposed to locate the referred target from a short description of visual space by bounding boxes~\cite{zhang-etal-2023-ck,subramanian-etal-2022-reclip,kim-etal-2022-flexible}, which is part of vision-language tasks~\cite{fei2024enhancing,fei2024video,fei2024vitron,gong2022person,wu24next,wu2024towards,ge2021structured,li2023transformer,li2024laso}.
Recent works in 3D-REC can be divided into two parts, two-stage and single-stage.
As for two-stage methods~\cite{Chen_Chang_Nießner_2020, Achlioptas_Abdelreheem_Xia_Elhoseiny_Guibas_2020, Zhao_Cai_Sheng_Xu_2021, Yuan_Yan_Liao_Zhang_Wang_Li_Cui_2021, Yang_Zhang_Wang_Luo_2021, Huang_Lee_Chen_Liu_2021, Feng_Li_Li_Zhang_Zhang_Zhu_Zhang_Wang_Mian_2021}, 3D object proposals are generated directly from ground-truth~\cite{Achlioptas_Abdelreheem_Xia_Elhoseiny_Guibas_2020} or extracted by a pre-trained 3D object detector~\cite{Qi_Litany_He_Guibas_2019} in the first stage, and then assigned to language in the second stage. 
In the other way, some methods adopt a one-stage paradigm~\cite{Luo_Fu_Kong_Gao_Ren_Shen_Xia_Liu_2022,Jain_Gkanatsios_Mediratta_Fragkiadaki_2021,Wu_Cheng_Zhang_Cheng_Zhang_2022,wu20243dgresgeneralized3dreferring}, enabling end-to-end training.

Referring Expression Segmentation (RES) need fine-grained vision-language alignment~\cite{li2022blip,li2023blip,ge20243shnet,li2022fine}, proposed to locate the referred target by masks~\cite{jain2022comprehensive,suo-etal-2023-text,huang-satoh-2023-referring}. TGNN~\cite{Huang_Lee_Chen_Liu_2021} introduce 3D-RES by extending the bounding box annotations of ScanRefer~\cite{Chen_Chang_Nießner_2020} to masks by incorporating the instance masks from ScanNet and proposed a two-stage pipeline.
Further, 3D-STMN~\cite{wu20233d} proposed an end-to-end method that matches the text and superpoints to get the 3D segmentation of the target object directly.

\subsection{3D Human-AI Interaction}
ScanQA~\cite{azuma2022scanqa} has notably advanced visual question answering in 3D scenes, enhancing the human-AI interaction experience. Meanwhile, 3D-LLM~\cite{hong20233d}, 3D-VisTA~\cite{zhu20233d}, NaviLLM~\cite{zheng2023towards}, and BridgeQA~\cite{mo2024bridging} have further propelled this task. \citet{li2022toist,li2023understanding,lu2024scaneru} have explored how AI understands human instructions like gestures and language to locate targets. 3D-VisTA~\cite{zhu20233d} introduced a new paradigm for large-scale 3D vision-language pre-training, greatly enhancing AI's understanding of 3D vision-language and advancing various downstream tasks. Works like 3D-LLM~\cite{hong20233d}, Chat3D~\cite{wang2023chat,huang2023chat}, NaviLLM~\cite{zheng2023towards} and Scene-LLM~\cite{fu2024scene} have extended the capabilities of multimodal large language models to the 3D realm, endowing embodied intelligence with the rich knowledge and capabilities of LLMs, thus ushering in the era of large models in Human-AI Interaction.

\subsection{Weakly Supervision in Vision-and-Language}
In the field of Vision Language, weakly supervised ~\cite{lee2023weakly, liu2023referring, li2023fully, banerjee2020weaqa} have gained significant attention and great progress. These approaches aim to tackle the challenge of limited or incomplete annotations by leveraging alternative supervised data or weakly labeled data.
For weakly supervised visual question answering (VQA), \citet{kervadec2019weak} employ weak supervision in the form of object-word alignment as a pre-training task. \citet{trott2017interpretable} use object counts in images as weak supervision to guide VQA for counting-based questions. \citet{gokhale2020vqa} employ logical connective rules to augment training datasets for yes-no questions. Weakly supervision from captions has also been employed for visual grounding tasks~\cite{fang2020weak, mithun2019weakly, anne2017localizing} recently.
Especially, for RES, some methods~\cite{lee2023weakly, liu2023referring} localize the target object only using readily available image-text pairs.

\section{Method}
\label{sec:method}
In this section, we provide a comprehensive overview of the RG-SAN. The framework is illustrated in Fig.~\ref{fig:2}. First, the features of visual and linguistic modalities are extracted in parallel (Sec.~\ref{sec: feat}). Next, we demonstrate the process of TLM (Sec.~\ref{sec:tlm}). Finally, we outline the RWS and the training objectives (Sec.~\ref{sec:rws}).

\subsection{Feature Extraction}\label{sec: feat}
\subsubsection{Visual Encoding} 
Given a point cloud scene $\mathbf{P}_{cloud}\in \mathbb{R}^{\mathcal{N}_p\times (3+F)}$ with $\mathcal{N}_p$ points. Each point comes with 3D coordinates along with an \textit{F}-dimensional auxiliary feature that includes RGB, normal vectors, among others. We first employ a Sparse 3D U-Net~\cite{Graham_Engelcke_Maaten_2018} to extract point-wise features, represented as $\mathbf{{\hat P_{cloud}}}\in \mathbb{R}^{\mathcal{N}_p\times C_p}$. Then, we follow~\citet{Sun_Qing_Tan_Xu_2022} and \citet{wu20233d} to obtain $\mathcal{N}_s$ superpoints $\{\mathcal{K}_i\}^{\mathcal{N}_s}_{i=1}$~\cite{Landrieu_Simonovsky_2018} from the original point cloud.
Finally, we directly feed point-wise features $\mathbf{{\hat P_{cloud}}}$ into superpoint pooling layer based on $\{\mathcal{K}_i\}^{\mathcal{N}_s}_{i=1}$ to obtain the superpoint-level features $\mathbf{S}_p \in \mathbb{R}^{\mathcal{N}_s\times C_p}$.

\subsubsection{Linguistic Encoding}
Given a free-form plain text description of the target object, consisting of $\mathcal{N}_t$ words $\{c_i\}_{i=1}^{\mathcal{N}_t}$, we utilize a pre-trained MPNet model~\cite{song2020mpnet} to extract $C_t$-dimensional word-level embeddings, represented as $\mathbf{E}_0 \in \mathbb{R}^{\mathcal{N}_t\times C_t}$.

\begin{figure*}[!t]
\centering 
  \includegraphics[width=0.9\columnwidth]{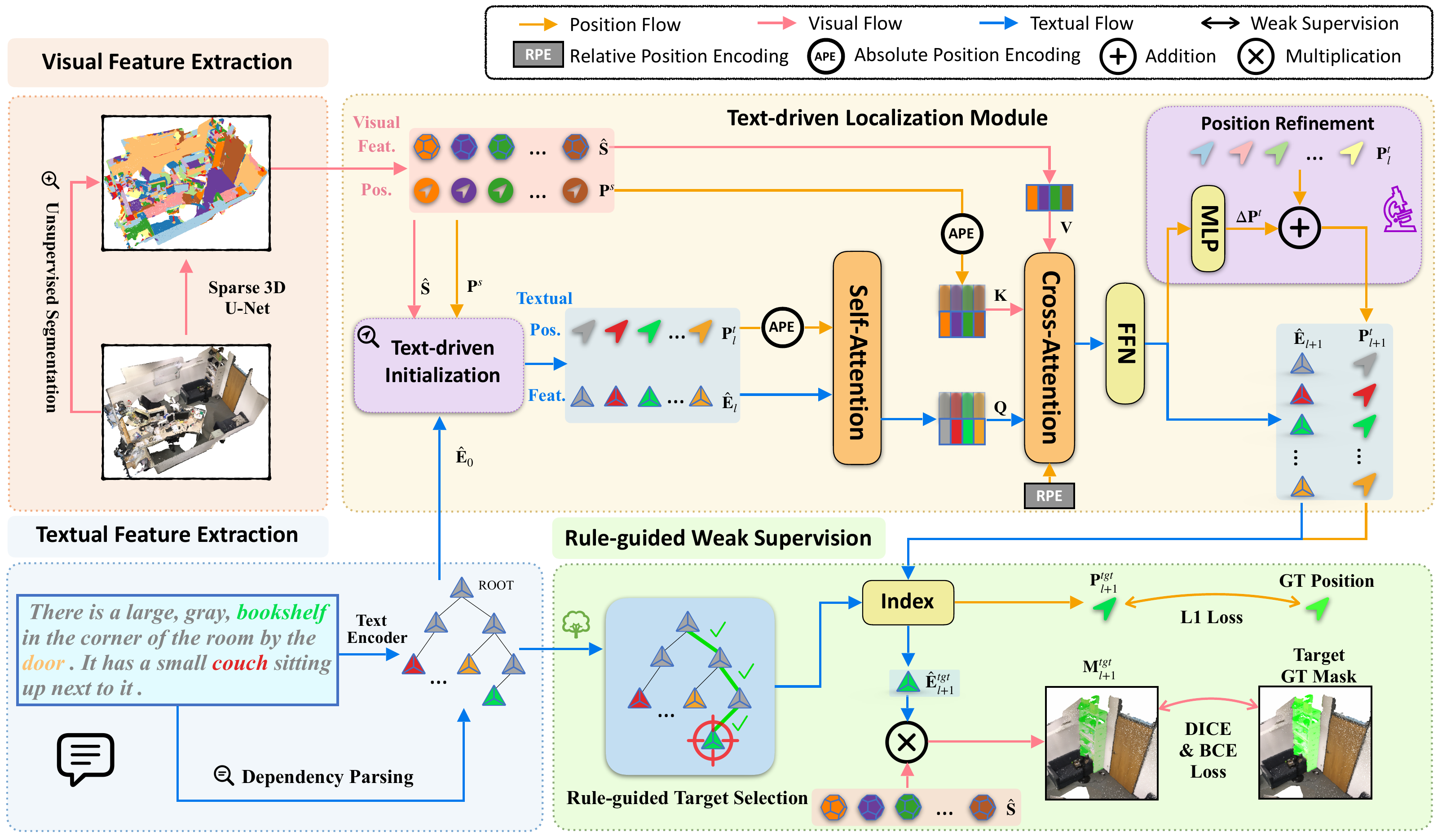}
  \caption{An overview of the proposed RG-SAN. This model analyzes a point cloud and a textual description with $\mathcal{N}_t$ tokens, extracting superpoints and word-level features. The TLM assigns spatial positions to tokens, facilitating multimodal fusion. The RWS strategy enables the model to learn the positions of all mentioned entities using only the supervision of the target position.
  }
  \label{fig:2}
\end{figure*}

\subsection{Context-driven Spatial Awareness}\label{sec:tlm}
In this section, we address a key limitation in prior works that interact point clouds with text without considering spatial positioning~\cite{wu20233d, Luo_Fu_Kong_Gao_Ren_Shen_Xia_Liu_2022, Wu_Cheng_Zhang_Cheng_Zhang_2022}. Unlike these methods, which often lose spatial information due to unordered point cloud features, leading to ambiguous spatial relationship understanding, our approach is distinct. In 3D-RES, spatial information is inherently sparse and dynamic, depending on the specific target object described in the text, rather than the dense, static sampling of an entire point cloud scene~\cite{lai2023mask}.

To address this issue, we propose to facilitate interactions between textual entities and point clouds within 3D space, rather than merely at the semantic level. Specifically, our objective is to fully leverage semantic and spatial contextual information to accurately predict the spatial positions of all mentioned nouns within the point cloud.

Therefore, we introduce the Text-driven Localization Module (TLM) to initialize the positions of entity nouns in the text and continuously update and refine these positions through iterative multimodal interactions.

\subsubsection{Text-driven Localization Module}\label{sec:tlm}
Given the superpoint features $\mathbf{S}_p$ and word embeddings $\mathbf{E}_0$, we first project the features into the same dimension, and enhance the word-level embeddings by Dependency-Driven Interaction (DDI), following~\citet{wu20233d}:
\begin{align}
    \hat{\mathbf{E}}_0 = \textbf{DDI}(\mathbf{E}_0\mathbf{W}_{lang}), \quad \hat{\mathbf{S}} = \mathbf{S}_p\mathbf{W}_{vis},
\end{align}
where $\mathbf{W}_{lang}\in\mathbb{R}^{C_t \times D}$ and $\mathbf{W}_{vis}\in \mathbb{R}^{C_p\times D}$ denote learnable parameters, and the subscript of $\mathbf{E}$ and $\hat{\mathbf{E}}$ represents the round number.

\noindent\textbf{Text-driven Initialization.}
The key is to map the text into 3D geometric space in a meaningful way.
Specifically, we enhance entity position prediction within point clouds through an interactive text-point cloud process. We do this by calculating feature similarity across modalities to accurately estimate the spatial probability distribution for each mentioned entity:
\begin{equation} \label{eq:general_attn}
    \begin{split}
        \mathbf{E}=\hat{\mathbf{E}}_0 \mathbf{W}_E, \quad \mathbf{S}=\hat{\mathbf{S}} \mathbf{W}_S,\\
        A_{ij}= \frac{{\rm Sim}{\left(\mathbf{E}_i,\mathbf{S}_j\right)}}{\sum_{j=1}^{\mathcal{N}_s}\ {\rm Sim}{\left(\mathbf{E}_i,\mathbf{S}_j\right)}},
    \end{split}
\end{equation}
where $\hat{\mathbf{E}}_0$ denotes the initial word embeddings, $\hat{\mathbf{S}}$ denotes the superpoint features, $\mathbf{W}_E,\mathbf{W}_S \in \mathbb{R}^{D \times D}$ are learnable parameters, $A_{ij}\in \mathbb{R}$ denotes the probability of the $i$-th word token being located at the $j$-th superpoint, and $ {\rm Sim}{\left(\cdot,\cdot \right)} $ represents the similarity function, which in this case is defined as $ {\rm Sim}\left(\mathbf{E},\mathbf{S}\right)\!=\!{\rm exp}({\mathbf{E}\mathbf{S}^T}/{\sqrt D}) $.

Following this, we utilize the spatial probability distribution $A$ to predict the approximate positions of the mentioned entities, as well as their corresponding representations:
\begin{equation}
\mathbf{P}^t_{0,i} = \sum_{j=1}^{\mathcal{N}_s}\ A_{ij} \mathbf{P}^s_j,
\end{equation}
\begin{equation}
    \begin{split}
    \mathbf{S}_v = \hat{\mathbf{S}}\mathbf{W}_{v},\quad
    \hat{\mathbf{E}}_{0,i} = \sum_{j=1}^{\mathcal{N}_s}\ A_{ij} \mathbf{S}_{v,j},
    \end{split}
\end{equation}
where $\mathbf{P}^s_j$ is the position of the $j$-th superpoint, $\mathbf{P}^t_{0,i}$ is the initial spatial position of $i$-th word token which will be refined iteratively as formulated in Sec.~\ref{sec:saa}, $\mathbf{W}_{v} \in \mathbb{R}^{D \times D}$ denotes learnable parameters, and $\hat{\mathbf{E}}_{0,i}$ denotes the updated representation of the $i$-th word token.
The sharing of distribution \( A \) during centroid computation allows the entity representations to benefit from the guidance provided by spatial information, leading to a more accurate understanding of the 3D spatial relationships. Subsequently, the text and point clouds undergo multiple rounds of multimodal interactions, continually updating the embeddings and positions of the entities.

\noindent\textbf{Iterative Position Refinement.}
After \( l \)-round multimodal interactions, the word tokens \( \hat{\mathbf{E}}_l \), referred to as textual segment kernels, become increasingly precise, theoretically resulting in more accurate position predictions. A straightforward approach would involve replicating the initial interaction method by regressing position information in each round. However, following the methodologies of \citet{redmon2016you} and \citet{lai2023mask}, rather than directly optimizing the final position, we adopt a more manageable strategy of iteratively learning offsets.
To this end, we refine the positions of textual tokens based on the evolving textual segment kernels. As depicted in Fig.~\ref{fig:2}, we employ a Multilayer Perceptron (MLP) to predict a position offset $\Delta \mathbf{P}^t_l = \textbf{MLP}(\hat{\mathbf{E}}_{l+1}) \in \mathbb{R}^{\mathcal{N}t \times 3}$ from the updated textual segment kernels $\hat{\mathbf{E}}_{l+1}$. This offset is then added to the previous textual positions $\mathbf{P}^t_l$: 
\begin{equation} 
\mathbf{P}^t_{l+1} = \mathbf{P}^t_l + \Delta \mathbf{P}^t_l.
\end{equation}
This method allows for gradual refinement of position predictions, making the optimization process more effective and leading to progressively more accurate positioning with each iteration.

\subsubsection{Spatial Awareness Aggregation}\label{sec:saa}
Once the positions of noun entities are obtained, techniques like positional encoding~\cite{vaswani2017attention,wu2021rethinking,lai2023mask,lai2022stratified,chen2022language} can be used to further refine the positions. 

\noindent\textbf{Absolute Positional Encoding (APE).} To initiate, we follow the approach of the original transformer~\cite{vaswani2017attention} to encoded the positions of both superpoints and text tokens to obtain positional encodings $\mathbf{B}^s_l \in \mathbb{R}^{\mathcal{N}_s\times D}$ and $\mathbf{B}^t_l \in \mathbb{R}^{\mathcal{N}_t\times D}$ using absolute positional encoding (APE):
\begin{equation}
\mathbf{B}^s_l = \textbf{APE}(\mathbf{P}^s_l), \quad \mathbf{B}^t_l = \textbf{APE}(\mathbf{P}^t_l).
\end{equation}
These positional encodings facilitate spatial-aware self-attention in the textural segment kernels $\hat{\mathbf{E}}_l$:
\begin{equation}\label{eq:ape_add}
\dot{\mathbf{E}}_l = \textbf{Attention}(\hat{\mathbf{E}}_l + \mathbf{B}^t_l, \hat{\mathbf{E}}_l + \mathbf{B}^t_l, \hat{\mathbf{E}}_l),
\end{equation}
where $\text{Attention}(\cdot)$ uses the technique of \citet{vaswani2017attention} and $\mathbf{B}^t_l$ denotes the absolute positional encoding of $\hat{\mathbf{E}}_l$.

Next, we enhance textual and superpoint features with absolute positional encoding, and use them as Queries and Keys for subsequent multimodal aggregation:
\begin{equation}\label{eq:rpe_add}
\begin{split}
    \mathbf{Q}&=\text{Concat}(\dot{\mathbf{E}}_{l},\mathbf{B}^t_l) \mathbf{W}_{query}, \\
    \mathbf{K}&=\text{Concat}(\hat{\mathbf{S}}, \mathbf{B}^s_l) \mathbf{W}_{key},
\end{split}
\end{equation}
where $\mathbf{B}^t_l \in \mathbb{R}^{\mathcal{N}_t\times D},\mathbf{B}^s_l \in \mathbb{R}^{\mathcal{N}_s\times D}$ denote the absolute positional encoding of segmentation kernels and superpoints, respectively, and $\mathbf{W}_{query},\mathbf{W}_{key}\in \mathbb{R}^{2D\times 2D}$ denote learnable parameters.

\noindent\textbf{Relative positional encoding (RPE).} 
For the further interaction with superpoint features, we adopt well-established relative positional encoding techniques~\cite{wu2021rethinking,lai2023mask,lai2022stratified,chen2022language}, such as Table-based RPE~\cite{wu2021rethinking,lai2023mask} and 5D Euclidean RPE~\cite{chen2022language}, which are formalized as follows:
\begin{equation}
{\mathbf{B}^r_l}{[i,j]} = \text{RPE}(\mathbf{Q}[i]+\mathbf{K}[j]),
\end{equation}
where $\mathbf{B}^r_l[i,j]\in \mathbb{R}$ denotes the relative positional bias of the $i$-th $\mathbf{Q}$ relative to the $j$-th $\mathbf{K}$, $\text{RPE}(\cdot)$ denotes the operation of relative positional bias and $[\cdot]$ denotes the indexing operation.

Thus, we can perform multimodal aggregation enhanced with relative positional encoding:
\begin{equation}
\hat{\mathbf{E}}_{l+1} = \text{softmax}\left(\frac{\mathbf{Q} \cdot \mathbf{K}^T}{\sqrt{D}}+\mathbf{B}^r_l\right) \cdot (\hat{\mathbf{S}}\mathbf{W}_{val}),
\end{equation}
where $\mathbf{W}_{val}\in \mathbb{R}^{D\times D}$ denote learnable parameters, $\mathbf{B}^r_l\in\mathbb{R}^{\mathcal{N}_t\times\mathcal{N}_s}$ denotes the relative positional bias, and $\hat{\mathbf{E}}_{l+1}$ denotes the updated segmentation kernels.
This methodology significantly enriches the interaction between linguistic and 3D visual data, enabling more nuanced spatial understanding in our model.

\subsection{Rule-guided Weak Supervision}\label{sec:rws}
\subsubsection{Rule-guided Target Selection}
In the preceding sections, we initially predicted the locations of all entities mentioned in the text. Ideally, supervised training would require position labels for each entity. However, we only have access to the location information of the target instance. This constraint leads us to adopt a weak supervision approach, focusing solely on the position of the referring instance for training. 
This approach introduces a significant challenge: accurately identifying the referring instance among the mentioned nouns. 
To address this, we utilize a pre-processed dependency tree, as outlined in~\citet{Manning_Surdeanu_Bauer_Finkel_Bethard_McClosky_2014}, to accurately pinpoint the core noun, typically the subject of the sentence. 
We have developed a set of manual rules, based on this more general dependency tree, to enhance the identification process. These rules are specifically designed to guide the accurate positioning of core instances. The implementation of these rules is outlined in Algorithm~\ref{alg:1}.
\begin{algorithm} 
\renewcommand{\algorithmicrequire}{\textbf{Input:}}
\renewcommand{\algorithmicensure}{\textbf{Output:}}
	\caption{Rule-guided Target Selection} 
	\label{alg3} 
	\begin{algorithmic}[1]
            \REQUIRE The dependency tree $\mathcal{G}=(\mathcal{V}, \mathcal{E})$ of the textual description, where $\mathcal{V}=\{\text{token}\}$ denotes the set of nodes, $\mathcal{E}=\{(\text{relation, head, tail})\}$ denotes the set of relations between nodes.
            \ENSURE The index $i$ of Target Instance node $\mathcal{V}^{tgt}$
            \STATE Initialization $i$ to the root: $i=0$
            \STATE find $\mathcal{E}_i$ with $\mathcal{V}_i$ as its head
            \IF{$(\mathcal{E}_i\in\{\text{nsubj, compound}\}) \And (\mathcal{V}_i \notin \{\text{which, that}\}) $}
                \STATE $i \gets \mathcal{E}_i$'s tail index
            \ENDIF
            \IF{$\mathcal{V}_i \in \{\text{there, this, it, object}\}$}
                \STATE find $\mathcal{E}_i$ with $\mathcal{V}_i$ as its head
                \STATE $i \gets \mathcal{E}_i$'s tail index
            \ENDIF
            \IF{$\mathcal{V}_i \in \{\text{set, sets, color, shape}\}$}
                \STATE find the first $\mathcal{E}_i\text{'s relation}\in \{\text{compound, nmod, dep}\}$
                \STATE $i \gets \mathcal{E}_i$'s head index
            \ENDIF
	\end{algorithmic} \label{alg:1}
\end{algorithm}

\subsubsection{Training Objectives}
Given the index of the target instance, we can directly obtain the corresponding segment kernel $\hat{\mathbf{E}}^{tgt}_{l+1}\in \mathbb{R}^{D}$ and position $\mathbf{P}^{tgt}_{l+1}$, which are then supervised by the target ground truth.

Then we perform matrix multiplication between $\hat{\mathbf{E}}^{tgt}_{l+1}$ and $\hat{\mathbf{S}}$ to get the predicted instance response maps, which can be formulated as
\begin{align}
\mathbf{M}_{l+1} &= \sigma(\hat{\mathbf{E}}^{tgt}_{l+1}\cdot\hat{\textbf{S}}^T),\\
\textbf{Mask}_{l+1} &= \mathbf{M}_{l+1}>0.5,
\end{align}
where $\mathbf{M}_{l+1}\in \mathbb{R}^{\mathcal{N}_s}$, $\textbf{Mask}_{l+1}\in \{0,1\}^{\mathcal{N}_s}$ are the predicted response map and the instance mask corresponding to the target.

Given ground-truth binary mask of the referring expression $\mathbf{Y} \in \{0,1\}^{\mathcal{N}_p}$, we get the corresponding superpoint mask $\mathbf{Y}^s \in \{0,1\}^{\mathcal{N}_s}$ by superpoint pooling follewed by a 0.5-threshold binarization, and then we apply the binary cross-entropy (BCE) loss on the final response map $\mathbf{M}_{l+1}$ following~\citet{Sun_Qing_Tan_Xu_2022}. The operation can be written as:
\begin{align}
        \mathbf{Y}^s_i &= \mathbb{I}(\sigma(\text{AvgPool}(\mathbf{Y},\mathcal{K}_i))),\\
        \mathcal{L}_{bce} &=  \text{BCE}(\mathbf{M}_{l+1},\mathbf{Y}^s),
\end{align}
where $\text{AvgPool}(\cdot)$ denotes the superpoint average pooling operation, and $\mathbf{Y}^s_i$ denotes the binarized mask value of the $i$-th superpoint $\mathcal{K}_i$. $\mathbb{I}(\cdot)$ indicates whether the mask value is higher than 50\%.

To tackle foreground-background sample imbalance, we can use Dice loss~\cite{Milletari_Navab_Ahmadi_2016}:
\begin{equation}
    \mathcal{L}_{dice} =  \text{DICE}(\mathbf{M}_{l+1},\mathbf{Y}^s).
\end{equation}

To supervise the position $\mathbf{P}^{tgt}_{l+1}$, we use the center of the superpoints of the target instance $\mathbf{P}^{gt}$, as
\begin{align}\label{eqn:pos}
    \mathcal{L}_{pos}=\text{L1}(\mathbf{P}^{tgt}_{l+1}, \mathbf{P}^{gt}).
\end{align}

In addition, we add a simple auxiliary score loss $\mathcal{L}_{score}$ for mask quality prediction following~\citet{Sun_Qing_Tan_Xu_2022}.

Overall, the final training loss function $\mathcal{L}$ can be formulated as:
\begin{equation}
\begin{split}
    \mathcal{L} &= \lambda_{bce}\mathcal{L}_{bce} + \lambda_{dice}\mathcal{L}_{dice} + \lambda_{pos}\mathcal{L}_{pos} 
    + \\
    &\lambda_{score}\mathcal{L}_{score},
\end{split}
\end{equation}
where $\lambda_{bce}$, $\lambda_{dice}$, $\lambda_{rel}$ and $\lambda_{score}$ are hyperparameters used to balance these four losses.

\section{Expriment}\label{sec:exp}
\subsection{Experiment Settings}
In our experiment, we utilize the pre-trained Sparse 3D U-Net method to extract point-wise features from point clouds~\cite{Sun_Qing_Tan_Xu_2022}. We also employ the pre-trained MPNet model~\cite{song2020mpnet} as our text encoder. For the rest of the network, training is conducted from scratch. We set an initial learning rate of 0.0001 and apply a learning rate decay at epochs 26, 34, and 46, each with a decay rate of 0.5. Our experiments use a default of 6 multiple rounds $L$, a batch size of 32, and a maximum sentence length of 80. We set $\lambda_{bce}=\lambda_{dice}=1, \lambda_{pos}=\lambda_{score}=0.5$. All experiments are conducted using PyTorch on a single NVIDIA Tesla A100 GPU, ensuring consistency in our computational process.

\begin{table*}\small
    \centering
    \caption{The 3D-RES results on ScanRefer. \dag~The mIoU and accuracy are reevaluated on our machine. $^*$We reproduce results by extracting points within the boxes as segmentation mask predictions using their official codes.}
    \resizebox{1\textwidth}{!}{
        \begin{tabular}{c|ccc|ccc|ccc|ccc}
            \toprule
            & \multicolumn{3}{c|}{Unique ($\sim$19\%)} 
            & \multicolumn{3}{c|}{Multiple ($\sim$81\%)} 
            & \multicolumn{3}{c|}{\textbf{Overall}}
            & \multicolumn{3}{c}{Inference Time} \\
            \multirow{-2}{*}{Method} 
            & 0.25 & 0.5 & mIoU 
            & 0.25 & 0.5 & mIoU 
            & \textbf{0.25} & \textbf{0.5} & \textbf{mIoU} 
            & Stage-1 & Stage-2 & All \\
            \midrule

            \multicolumn{10}{l}{{   Multi-task}}\\
            \midrule

            EDA-box2mask~\cite{Wu_Cheng_Zhang_Cheng_Zhang_2022}  
            & {84.7} & {56.9} & - 
            & {50.0} & {37.0} & - 
            & {55.2} & {40.0} & {35.0} 
            & - & - & - \\

            3DRefTR-SP~\cite{lin2023unified}  
            & {87.9} & {69.8} & - 
            & {51.6} & {41.9} & - 
            & {57.0} & {46.1} & {40.8} 
            & - & - & 388ms \\

            3DRefTR-HR~\cite{lin2023unified}  
            & {89.6} &{77.0} & - 
            & {52.3} & {43.7} & - 
            & {57.9} & {48.7} & {41.2} 
            & - & - & 405ms \\

            UniSeg3D~\cite{xu2024unified}
            & - & - & - 
            & - & - & - 
            & - & - & 29.6
            & - & - & - \\
            
            SegPoint~\cite{he2025segpoint}
            & - & - & - 
            & - & - & - 
            & - & - & 41.7
            & - & - & - \\

            Reason3D~\cite{huang2024reason3d}
            & 88.4 & 84.2 & 74.6 
            & 50.5 & 31.7 & 34.1 
            & 57.9 & 41.9 & 42.0
            & - & - & - \\
            
            \midrule

            \multicolumn{10}{l}{{ Single-task}}\\
            \midrule

            TGNN~\cite{Huang_Lee_Chen_Liu_2021} 
            & - & - & - 
            & - & - & - 
            & 37.5 & 31.4 & 27.8 
            & - & - & - \\  

            TGNN\dag~\cite{Huang_Lee_Chen_Liu_2021} 
            & {69.3} & {57.8} & {50.7} 
            & {31.2} & {26.6} & {23.6} 
            & {38.6} & {32.7} & {28.8} 
            & 26862ms & 235ms & 27097ms\\   
            
            InstanceRefer\dag~\cite{Yuan_Yan_Liao_Zhang_Wang_Li_Cui_2021} 
            & 81.6 & 72.2 & 60.4 
            & 29.4 & 23.5 & 21.5 
            & 40.2 & 33.5 & 30.6 
            & 509ms & 672ms & 1181ms \\

            X-RefSeg3D~\cite{qian2024x}
            & - & - & - 
            & - & - & - 
            & 40.3 & 33.8 & 29.9
            & - & - & - \\

            3DVG-Transformer$^*$~\cite{Zhao_Cai_Sheng_Xu_2021} 
            & 79.5 & 58.0 & 49.9 
            & 42.0 & 30.8 & 27.0 
            & 49.3 & 36.1 & 31.4
            & - & - & - \\

            3D-SPS$^*$~\cite{Luo_Fu_Kong_Gao_Ren_Shen_Xia_Liu_2022}
            & 84.8 & 65.6 & 54.7 
            & 41.7 & 30.8 & 26.7 
            & 50.1 & 37.6 & 32.1
            & - & - & - \\
            
            3DRESTR~\cite{lin2023unified}  
            & {79.0} & {54.2} & - 
            & {40.2} & {22.1} & - 
            & {46.0} & {26.9} & {28.7} 
            & - & - & - \\

            3D-STMN~\cite{wu20233d}  
            & \textbf{89.3} & {84.0} & \textbf{74.5} 
            & {46.2} & {29.2} & {31.1} 
            & {54.6} & {39.8} & {39.5}  
            & - & - & 283ms \\ 

            RG-SAN (Ours)  
            & {89.2} & \textbf{84.3} & \textbf{74.5} 
            & \textbf{55.0} & \textbf{35.4} & \textbf{37.4} 
            & \textbf{61.7} & \textbf{44.9} & \textbf{44.6} 
            & - & - & 295ms \\
            \bottomrule
        \end{tabular}
    }
    \label{tab:scanrefer_benchmark}
\end{table*}

\subsection{Dataset and Evaluation Metrics}
We evaluate our method using the ScanRefer dataset, a recent 3D referring dataset~\cite{Chen_Chang_Nießner_2020, Huang_Lee_Chen_Liu_2021}, comprising 51,583 English natural language expressions referring to 11,046 objects across 800 ScanNet scenes~\cite{Dai_Chang_Savva_Halber_Funkhouser_Niessner_2017}. 
Following \citet{Chen_Chang_Nießner_2020}, our evaluation metrics include mean Intersection over Union (mIoU) and Acc@$k$IoU. ``Unique'' refers to cases where the target instance is the only one of its class, and ``Multiple'' indicates situations where there is at least one more object of the target's class.

\subsection{Quantitative Comparison}

In our experiments on the ScanRefer dataset, our proposed RG-SAN demonstrates significant improvements in nearly all metrics on the single-task leaderboard, as shown in Tab.~\ref{tab:scanrefer_benchmark}. Notably, RG-SAN shows substantial gains compared to the state-of-the-art single-task model 3D-STMN, with increases of 5.1 points in mIoU and 7.1 points in Acc@0.25. This highlights our model's inferencing capability. 
A more detailed examination reveals that the majority of these improvements occur in scenarios with multiple disruptive instances, where RG-SAN achieves a remarkable 6.3-point increase in mIoU. This setting, where the target instance is among other instances of the same type, demands discriminative reasoning from the model. The significant performance validates the enhanced referring capabilities empowered by spatial reasoning. Our proposed RG-SAN also outperforms multi-task models~\cite{Wu_Cheng_Zhang_Cheng_Zhang_2022,lin2023unified}, including LLM-based models~\cite{he2025segpoint,huang2024reason3d}, in most 3D-RES metrics, despite those models benefiting from more annotated data.

Moreover, RG-SAN has competitive inference costs, being only 12ms slower than the efficient 3D-STMN and faster than all other compared models, demonstrating its high performance with minimal computational increase.

\subsection{Ablation Study}
\subsubsection{Text-driven Localization Module}

We conduct an ablation study on the Text-driven Localization Module (TLM), as illustrated in Tab.~\ref{tab:query_init}. Simultaneously, we perform a fine-grained analysis of various initialization schemes for embeddings and positions. The term "w/o TLM" denotes the approach of not modeling positional information and instead directly using text embeddings for interaction. "MAFT" refers to the direct adaptation of the method proposed in~\cite{lai2023mask}. The "Project" method involves initializing embeddings based on text-driven embeddings and then projecting each textual token directly into a 3D position, while the "Random" method randomly assigns a position to each textual token. Finally, we utilize the initialization technique called Text-driven Initialization (TI), which simultaneously initializes both embeddings and positions in a text-driven manner.
Tab.~\ref{tab:query_init} clearly shows that, under identical conditions, TI outperforms the others in all metrics. This indicates that TI more effectively leverages positional information from the visual scene, leading to more precise initial positions for the textual tokens. Consequently, this reduces the complexity of the subsequent iterative refinement process, thereby enhancing the overall accuracy of our model in spatially aligning text with point cloud data. Additionally, Tab.~\ref{tab:query_init} demonstrates that proper initialization leads to the superior performance of TLM compared to the methods without TLM.

\begin{table}
\centering
\begin{minipage}{0.47\textwidth}
\caption{Ablation study of Text-driven Localization Module (TLM), where ``w/o TLM'' means not using TLM.}
\centering
\resizebox{1.0\columnwidth}{!}{
\begin{tabular}{c|cc|c|c}
\toprule
            & Init. of & Init. of & \multicolumn{1}{c|}{Multiple} 
            & \multicolumn{1}{c}{\textbf{Overall}}
            \\
            \multirow{-2}{*}{Method} 
            & Embeddings
            & Positions
            & mIoU 
            & \textbf{mIoU} \\

\midrule
w/o TLM & Text-driven & - & 32.5 & 40.3 \\
MAFT~\cite{lai2023mask} & Zero & Random & 29.7 & 37.9\\
\midrule
Project & Text-driven & Project & 30.3 & {38.8} \\  
Random & Text-driven & Random & 30.1 & {38.8} \\  
TI & Text-driven & Text-driven & \textbf{34.1} & \textbf{42.3} \\ 
\bottomrule
\end{tabular}
}
\label{tab:query_init}
\end{minipage}
 \hfill
\begin{minipage}{0.47\textwidth}
\caption{Ablation study of positional encoding, where ``w/o Pos. Supervision'' means not supervising the positions, and ``w/o PE'' means not using any positional encoding.}
\label{tab:pe}
\centering
\resizebox{1.0\columnwidth}{!}{
\begin{tabular}{c|ccc|ccc}
\toprule
& \multicolumn{3}{c|}{Multiple} 
            & \multicolumn{3}{c}{\textbf{Overall}}
            \\
            \multirow{-2}{*}{positional encoding} 
            & 0.25 & 0.5 & mIoU 
            & \textbf{0.25} & \textbf{0.5} & \textbf{mIoU} \\
\midrule

w/o Pos. Supervision & 45.4 & 27.3 & 30.4 & 54.4 & 38.2 & 38.9 \\

\midrule

w/o PE  & 46.1 & 31.7 & 32.8 & {54.6} & {42.4} & {41.1} \\  

Fourier APE & 46.0 & 30.9 & 32.0 & {55.1} & {41.5} & {40.7} \\

5D Euclidean RPE & 46.7 & 32.5 & 33.3 & {54.6} & {43.9} & {41.7} \\

Table-based RPE & \textbf{47.2} & \textbf{33.7} & \textbf{34.1}  & \textbf{55.6} & \textbf{43.9} & \textbf{42.3} \\

\bottomrule
\end{tabular}
}
\end{minipage}
\end{table}

\subsubsection{Positional Encoding}
We compare various positional encoding methods previously employed in~\cite{schult2022mask3d,chen2022language,lai2023mask}. These methods include Fourier Absolute positional encoding (APE), 5D Euclidean Relative positional encoding (5D Euclidean RPE)~\cite{chen2022language}, and Table-based Relative positional encoding (Table-based RPE)~\cite{lai2023mask}. Tab.~\ref{tab:pe} reveals that Table-based RPE surpasses the other methods, suggesting that combining semantic information with relative relationships is advantageous. Additionally, we observe that employing only absolute positional encoding can result in lower performance than not using any positional encoding at all. This may be attributed to the inherent limitations of absolute positional encoding in capturing relative positional information. By complicating the semantic features, it introduces challenges in the model's training process, underscoring the importance of choosing the right positional encoding technique for effective performance.

\subsubsection{Rule-guided Weak Supervision}
We conducted experiments employing various weakly supervised text kernel selection strategies to evaluate their efficacy in leveraging target annotations. The strategy labeled as "w/o RWS" involves selecting the token based on attention weight within the cross-attention module~\cite{wu20233d}, while "Root" entails selecting the root token of the dependency tree. Table~\ref{tab:wss} illustrates that utilizing the root node as supervision slightly outperforms the "w/o RWS" baseline. This is likely due to the root node providing consistent supervision, whereas Top1 tends to select different nodes variably, which complicates the training process.
In contrast, our Rule-guided Target Selection (RTS) strategy, based on dependency tree rules to locate subjects, aligns more effectively with the structural nature of the text. It precisely identifies the target entity's position, significantly enhancing annotation utilization and effectively directing model training. This leads to a notable improvement in model performance.

Furthermore, we conduct an ablation study on the impact of the position loss weight $\mathcal{L}_{pos}$, detailed in Tab.~\ref{tab:loss}. We observe that increasing the weight generally improves performance, peaking at a weight of 0.5, beyond which performance begins to taper off. This finding highlights the importance of balancing the weight of the position loss to optimize the model's effectiveness.

\subsubsection{Comparison with MAFT}
MAFT~\cite{lai2023mask} has played a pivotal role in 3D instance segmentation by incorporating spatial position modeling, offering valuable insights into how spatial information can improve model performance. Inspired by this approach, we extend spatial information into the text space to better align visual and textual semantics, specifically targeting spatial relationship reasoning in 3D-RES. Our approach introduces two key innovations that distinguish it from MAFT:
\begin{enumerate}[0]
    \item[$\bullet$] Unlike MAFT~\cite{lai2023mask}, which initializes queries with zeros and uses random initialization for positional information, we employ text-driven queries and positional information to model the spatial relationships of entities in the expressions. This allows our model to capture the spatial context better, resulting in a 4.4-point improvement in mIoU, as shown in Tab.~\ref{tab:query_init}
    \item[$\bullet$] In contrast to~\cite{lai2023mask}, which supervises the positions of all target instances, 3D-RES supervises only the core target word. Our novel RWS method constructs spatial relationships for all noun instances using only the target word's positional information, resulting in a 2.3-point improvement in mIoU, as demonstrated in Tab.~\ref{tab:wss}.
\end{enumerate}

\begin{table}
\begin{minipage}{0.47\textwidth}
\caption{Weak Supervision Strategy in RWS, where ``w/o RWS'' means using attention-based Top1 approach in \cite{wu20233d} instead of our RWS, and ``RTS'' refers to our Rule-guided Target Selection strategy.}
\centering
\resizebox{1.0\columnwidth}{!}{
\begin{tabular}{c|ccc|ccc}
\toprule
& \multicolumn{3}{c|}{Multiple} 
& \multicolumn{3}{c}{\textbf{Overall}}
\\
\multirow{-2}{*}{Strategy} 
& 0.25 & 0.5 & mIoU 
& \textbf{0.25} & \textbf{0.5} & \textbf{mIoU} \\
\midrule

w/o RWS & 47.2 & 33.7 & 34.1 & 55.6 & 43.9 & 42.3 \\  
\midrule
Root & 53.5 & 30.4 & 34.7 & 60.7 & 40.9 & 42.5 \\


RTS & \textbf{55.0} & \textbf{35.4} & \textbf{37.4}  & \textbf{61.7} & \textbf{44.9} & \textbf{44.6} \\

\bottomrule
\end{tabular}
}
\label{tab:wss}
\end{minipage}
 \hfill
\begin{minipage}{0.47\textwidth}
\caption{Ablation study of the weight of $\mathcal{L}_{pos}$.}
\centering
\resizebox{1.0\columnwidth}{!}{
\begin{tabular}{c|ccc|ccc}
\toprule
& \multicolumn{3}{c|}{Multiple} 
& \multicolumn{3}{c}{\textbf{Overall}}
\\
\multirow{-2}{*}{Weight of $\mathcal{L}_{pos}$} 
& 0.25 & 0.5 & mIoU 
& \textbf{0.25} & \textbf{0.5} & \textbf{mIoU} \\
\midrule

0.1 & 54.7 & 34.9 & 36.9 & 61.3 & 44.3 & 44.0 \\  

0.2 & \textbf{55.5} & 34.0 & 37.0 & \textbf{62.0} & 43.7 & 44.2 \\

0.5 & {55.0} & \textbf{35.4} & \textbf{37.4} & {61.7} & \textbf{44.9} & \textbf{44.6} \\

1.0 & 55.3 & 34.5 & 37.0 & 61.8 & 44.2 & 44.3 \\

2.0 & 54.3 & 33.8 & 36.6 & 61.0 & 43.7 & 43.9 \\

\bottomrule
\end{tabular}
}
\label{tab:loss}
\end{minipage}
\end{table}

\begin{figure*}
    \centering
    \includegraphics[width=1.0\columnwidth]{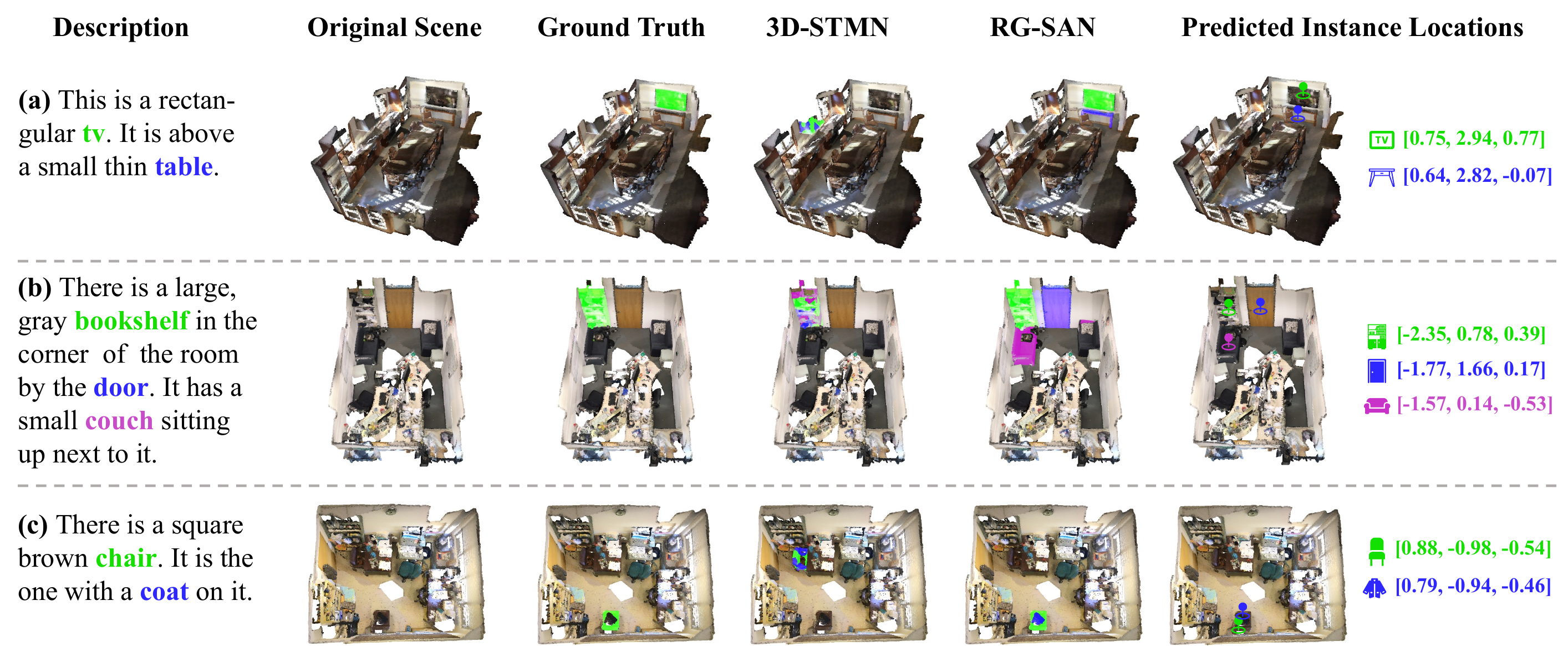}
    \caption{Visualization of all the nouns in the textual description. Our RG-SAN can segment instances corresponding to different nouns, while 3D-STMN indiscriminately assigns all nouns to the target instance. \textbf{Zoom in for best view.}}
    \label{fig:4}
\end{figure*}

\subsection{Qualitative Comparison}
We conduct a qualitative analysis on the ScanRefer validation set as shown in Fig.~\ref{fig:4}, comparing our proposed RG-SAN with 3D-STMN~\cite{wu20233d} to highlight our model's exceptional referring capability. Fig.~\ref{fig:4} demonstrates our model's ability to accurately segment not only the target objects but also other nouns mentioned in the text. Unlike 3D-STMN, which misattributes all nouns to a single target, RG-SAN distinctly recognizes and locates each noun. For example, in Fig.~\ref{fig:4}-\textbf{(c)}, our model successfully identifies the target chair through relative positioning, even with similar objects in the scene, and accurately recognizes a coat as a supporting element in the description. This ability extends to Fig.~\ref{fig:4}-\textbf{(a)} and \textbf{(b)}, where RG-SAN correctly segments multiple auxiliary nouns into their corresponding instances, demonstrating its robust generalization for complex texts and precise localization for multiple entities. Such capabilities enhance the model's understanding of complex semantic scenes, significantly improving its ability to refer to specific entities accurately.

\section{Conclusion}
In this paper, we present RG-SAN to overcome the limitations of traditional 3D-RES methods, particularly their lack of spatial awareness. Specifically, the TLM is introduced to model and refine positional information, while the RWS is designed to employ dependency tree rules to accurately guide the position of the target object. Combining TLM with RWS strategy, RG-SAN significantly improves segmentation accuracy and robustly handles spatial ambiguities. Extensive experiments conducted on the ScanRefer benchmark demonstrate the superior performance of RG-SAN. This underscores the importance of incorporating spatial awareness into segmentation models, paving the way for future advancements in the domain.

\section{Acknowledge}
This work was supported by National Science and Technology Major Project (No. 2022ZD0118201), the National Science Fund for Distinguished Young Scholars (No.62025603), the National Natural Science Foundation of China (No. U22B2051, No. U21B2037, No. 62072389, No. 62302411, No. 623B2088), the Natural Science Foundation of Fujian Province of China (No.2021J06003) and China Postdoctoral Science Foundation (No. 2023M732948).

\clearpage
\bibliographystyle{plainnat} 
{
	\small
	\bibliography{egbib}

\begin{thebibliography}{77}
\providecommand{\natexlab}[1]{#1}
\providecommand{\url}[1]{\texttt{#1}}
\expandafter\ifx\csname urlstyle\endcsname\relax
  \providecommand{\doi}[1]{doi: #1}\else
  \providecommand{\doi}{doi: \begingroup \urlstyle{rm}\Url}\fi

\bibitem[Achlioptas et~al.(2020)Achlioptas, Abdelreheem, Xia, Elhoseiny, and Guibas]{Achlioptas_Abdelreheem_Xia_Elhoseiny_Guibas_2020}
Panos Achlioptas, Ahmed Abdelreheem, Fei Xia, Mohamed Elhoseiny, and Leonidas Guibas.
\newblock Referit3d: Neural listeners for fine-grained 3d object identification in real-world scenes.
\newblock In \emph{Computer Vision--ECCV 2020: 16th European Conference, Glasgow, UK, August 23--28, 2020, Proceedings, Part I 16}, pages 422--440. Springer, 2020.

\bibitem[Anne~Hendricks et~al.(2017)Anne~Hendricks, Wang, Shechtman, Sivic, Darrell, and Russell]{anne2017localizing}
Lisa Anne~Hendricks, Oliver Wang, Eli Shechtman, Josef Sivic, Trevor Darrell, and Bryan Russell.
\newblock Localizing moments in video with natural language.
\newblock In \emph{Proceedings of the IEEE international conference on computer vision}, pages 5803--5812, 2017.

\bibitem[Azuma et~al.(2022)Azuma, Miyanishi, Kurita, and Kawanabe]{azuma2022scanqa}
Daichi Azuma, Taiki Miyanishi, Shuhei Kurita, and Motoaki Kawanabe.
\newblock Scanqa: 3d question answering for spatial scene understanding.
\newblock In \emph{proceedings of the IEEE/CVF conference on computer vision and pattern recognition}, pages 19129--19139, 2022.

\bibitem[Banerjee et~al.(2020)Banerjee, Gokhale, Yang, and Baral]{banerjee2020weaqa}
Pratyay Banerjee, Tejas Gokhale, Yezhou Yang, and Chitta Baral.
\newblock Weaqa: Weak supervision via captions for visual question answering.
\newblock \emph{arXiv preprint arXiv:2012.02356}, 2020.

\bibitem[Chen et~al.(2020)Chen, Chang, and Nie{\ss}ner]{Chen_Chang_Nießner_2020}
Dave~Zhenyu Chen, Angel~X Chang, and Matthias Nie{\ss}ner.
\newblock Scanrefer: 3d object localization in rgb-d scans using natural language.
\newblock In \emph{European conference on computer vision}, pages 202--221. Springer, 2020.

\bibitem[Chen et~al.(2022)Chen, Guhur, Tapaswi, Schmid, and Laptev]{chen2022language}
Shizhe Chen, Pierre-Louis Guhur, Makarand Tapaswi, Cordelia Schmid, and Ivan Laptev.
\newblock Language conditioned spatial relation reasoning for 3d object grounding.
\newblock \emph{Advances in Neural Information Processing Systems}, 35:\penalty0 20522--20535, 2022.

\bibitem[Dai et~al.(2017)Dai, Chang, Savva, Halber, Funkhouser, and Nie{\ss}ner]{Dai_Chang_Savva_Halber_Funkhouser_Niessner_2017}
Angela Dai, Angel~X Chang, Manolis Savva, Maciej Halber, Thomas Funkhouser, and Matthias Nie{\ss}ner.
\newblock Scannet: Richly-annotated 3d reconstructions of indoor scenes.
\newblock In \emph{Proceedings of the IEEE conference on computer vision and pattern recognition}, pages 5828--5839, 2017.

\bibitem[Devlin et~al.(2018)Devlin, Chang, Lee, and Toutanova]{Devlin_Chang_Lee_Toutanova_2019}
Jacob Devlin, Ming-Wei Chang, Kenton Lee, and Kristina Toutanova.
\newblock Bert: Pre-training of deep bidirectional transformers for language understanding.
\newblock \emph{arXiv preprint arXiv:1810.04805}, 2018.

\bibitem[Fang et~al.(2020)Fang, Kong, Wang, Fowlkes, and Yang]{fang2020weak}
Zhiyuan Fang, Shu Kong, Zhe Wang, Charless Fowlkes, and Yezhou Yang.
\newblock Weak supervision and referring attention for temporal-textual association learning.
\newblock \emph{arXiv preprint arXiv:2006.11747}, 2020.

\bibitem[Fei et~al.(2024{\natexlab{a}})Fei, Wu, Ji, Zhang, Zhang, Lee, and Hsu]{fei2024video}
Hao Fei, Shengqiong Wu, Wei Ji, Hanwang Zhang, Meishan Zhang, Mong-Li Lee, and Wynne Hsu.
\newblock Video-of-thought: Step-by-step video reasoning from perception to cognition.
\newblock In \emph{Forty-first International Conference on Machine Learning}, 2024{\natexlab{a}}.

\bibitem[Fei et~al.(2024{\natexlab{b}})Fei, Wu, Zhang, Chua, and Yan]{fei2024vitron}
Hao Fei, Shengqiong Wu, Hanwang Zhang, Tat-Seng Chua, and Shuicheng Yan.
\newblock Vitron: A unified pixel-level vision llm for understanding, generating, segmenting, editing, 2024{\natexlab{b}}.

\bibitem[Fei et~al.(2024{\natexlab{c}})Fei, Wu, Zhang, Zhang, Chua, and Yan]{fei2024enhancing}
Hao Fei, Shengqiong Wu, Meishan Zhang, Min Zhang, Tat-Seng Chua, and Shuicheng Yan.
\newblock Enhancing video-language representations with structural spatio-temporal alignment.
\newblock \emph{IEEE Transactions on Pattern Analysis and Machine Intelligence}, 2024{\natexlab{c}}.

\bibitem[Feng et~al.(2021)Feng, Li, Li, Zhang, Zhang, Zhu, Zhang, Wang, and Mian]{Feng_Li_Li_Zhang_Zhang_Zhu_Zhang_Wang_Mian_2021}
Mingtao Feng, Zhen Li, Qi~Li, Liang Zhang, XiangDong Zhang, Guangming Zhu, Hui Zhang, Yaonan Wang, and Ajmal Mian.
\newblock Free-form description guided 3d visual graph network for object grounding in point cloud.
\newblock In \emph{Proceedings of the IEEE/CVF International Conference on Computer Vision}, pages 3722--3731, 2021.

\bibitem[Fu et~al.(2024)Fu, Liu, Chen, Nie, and Xiong]{fu2024scene}
Rao Fu, Jingyu Liu, Xilun Chen, Yixin Nie, and Wenhan Xiong.
\newblock Scene-llm: Extending language model for 3d visual understanding and reasoning.
\newblock \emph{arXiv preprint arXiv:2403.11401}, 2024.

\bibitem[Ge et~al.(2021)Ge, Chen, Jose, Ji, Wu, and Liu]{ge2021structured}
Xuri Ge, Fuhai Chen, Joemon~M Jose, Zhilong Ji, Zhongqin Wu, and Xiao Liu.
\newblock Structured multi-modal feature embedding and alignment for image-sentence retrieval.
\newblock In \emph{Proceedings of the 29th ACM international conference on multimedia}, pages 5185--5193, 2021.

\bibitem[Ge et~al.(2024)Ge, Xu, Chen, Wang, Wang, An, and Jose]{ge20243shnet}
Xuri Ge, Songpei Xu, Fuhai Chen, Jie Wang, Guoxin Wang, Shan An, and Joemon~M Jose.
\newblock 3shnet: Boosting image--sentence retrieval via visual semantic--spatial self-highlighting.
\newblock \emph{Information Processing \& Management}, 61\penalty0 (4):\penalty0 103716, 2024.

\bibitem[Gokhale et~al.(2020)Gokhale, Banerjee, Baral, and Yang]{gokhale2020vqa}
Tejas Gokhale, Pratyay Banerjee, Chitta Baral, and Yezhou Yang.
\newblock Vqa-lol: Visual question answering under the lens of logic.
\newblock In \emph{European conference on computer vision}, pages 379--396. Springer, 2020.

\bibitem[Gong et~al.(2022)Gong, Huang, and Chen]{gong2022person}
Yunpeng Gong, Liqing Huang, and Lifei Chen.
\newblock Person re-identification method based on color attack and joint defence.
\newblock In \emph{Proceedings of the IEEE/CVF conference on computer vision and pattern recognition}, pages 4313--4322, 2022.

\bibitem[Graham et~al.(2018)Graham, Engelcke, and Van Der~Maaten]{Graham_Engelcke_Maaten_2018}
Benjamin Graham, Martin Engelcke, and Laurens Van Der~Maaten.
\newblock 3d semantic segmentation with submanifold sparse convolutional networks.
\newblock In \emph{Proceedings of the IEEE conference on computer vision and pattern recognition}, pages 9224--9232, 2018.

\bibitem[He et~al.(2025)He, Ding, Jiang, and Wen]{he2025segpoint}
Shuting He, Henghui Ding, Xudong Jiang, and Bihan Wen.
\newblock Segpoint: Segment any point cloud via large language model.
\newblock In \emph{European Conference on Computer Vision}, pages 349--367. Springer, 2025.

\bibitem[Hong et~al.(2023)Hong, Zhen, Chen, Zheng, Du, Chen, and Gan]{hong20233d}
Yining Hong, Haoyu Zhen, Peihao Chen, Shuhong Zheng, Yilun Du, Zhenfang Chen, and Chuang Gan.
\newblock 3d-llm: Injecting the 3d world into large language models.
\newblock \emph{Advances in Neural Information Processing Systems}, 36:\penalty0 20482--20494, 2023.

\bibitem[Huang et~al.(2023)Huang, Wang, Huang, Liu, Cheng, Zhao, Jin, and Zhao]{huang2023chat}
Haifeng Huang, Zehan Wang, Rongjie Huang, Luping Liu, Xize Cheng, Yang Zhao, Tao Jin, and Zhou Zhao.
\newblock Chat-3d v2: Bridging 3d scene and large language models with object identifiers.
\newblock \emph{arXiv preprint arXiv:2312.08168}, 2023.

\bibitem[Huang et~al.(2024)Huang, Li, Qi, Yan, and Yang]{huang2024reason3d}
Kuan-Chih Huang, Xiangtai Li, Lu~Qi, Shuicheng Yan, and Ming-Hsuan Yang.
\newblock Reason3d: Searching and reasoning 3d segmentation via large language model.
\newblock \emph{arXiv preprint arXiv:2405.17427}, 2024.

\bibitem[Huang et~al.(2021)Huang, Lee, Chen, and Liu]{Huang_Lee_Chen_Liu_2021}
Pin-Hao Huang, Han-Hung Lee, Hwann-Tzong Chen, and Tyng-Luh Liu.
\newblock Text-guided graph neural networks for referring 3d instance segmentation.
\newblock In \emph{Proceedings of the AAAI Conference on Artificial Intelligence}, volume~35, pages 1610--1618, 2021.

\bibitem[Huang and Satoh(2023)]{huang-satoh-2023-referring}
Ziling Huang and Shin{'}ichi Satoh.
\newblock Referring image segmentation via joint mask contextual embedding learning and progressive alignment network.
\newblock In Houda Bouamor, Juan Pino, and Kalika Bali, editors, \emph{Proceedings of the 2023 Conference on Empirical Methods in Natural Language Processing}, pages 7753--7762, Singapore, December 2023. Association for Computational Linguistics.
\newblock \doi{10.18653/v1/2023.emnlp-main.481}.
\newblock URL \url{https://aclanthology.org/2023.emnlp-main.481}.

\bibitem[Jain et~al.(2022)Jain, Gkanatsios, Mediratta, and Fragkiadaki]{Jain_Gkanatsios_Mediratta_Fragkiadaki_2021}
Ayush Jain, Nikolaos Gkanatsios, Ishita Mediratta, and Katerina Fragkiadaki.
\newblock Bottom up top down detection transformers for language grounding in images and point clouds.
\newblock In \emph{European Conference on Computer Vision}, pages 417--433. Springer, 2022.

\bibitem[Jain and Gandhi(2022)]{jain2022comprehensive}
Kanishk Jain and Vineet Gandhi.
\newblock Comprehensive multi-modal interactions for referring image segmentation.
\newblock In \emph{Findings of the Association for Computational Linguistics: ACL 2022}, pages 3427--3435, 2022.

\bibitem[Kervadec et~al.(2019)Kervadec, Antipov, Baccouche, and Wolf]{kervadec2019weak}
Corentin Kervadec, Grigory Antipov, Moez Baccouche, and Christian Wolf.
\newblock Weak supervision helps emergence of word-object alignment and improves vision-language tasks.
\newblock \emph{arXiv preprint arXiv:1912.03063}, 2019.

\bibitem[Kim et~al.(2022)Kim, Chu, and Kurohashi]{kim-etal-2022-flexible}
Yongmin Kim, Chenhui Chu, and Sadao Kurohashi.
\newblock Flexible visual grounding.
\newblock In Samuel Louvan, Andrea Madotto, and Brielen Madureira, editors, \emph{Proceedings of the 60th Annual Meeting of the Association for Computational Linguistics: Student Research Workshop}, pages 285--299, Dublin, Ireland, May 2022. Association for Computational Linguistics.
\newblock \doi{10.18653/v1/2022.acl-srw.22}.
\newblock URL \url{https://aclanthology.org/2022.acl-srw.22}.

\bibitem[Lai et~al.(2022)Lai, Liu, Jiang, Wang, Zhao, Liu, Qi, and Jia]{lai2022stratified}
Xin Lai, Jianhui Liu, Li~Jiang, Liwei Wang, Hengshuang Zhao, Shu Liu, Xiaojuan Qi, and Jiaya Jia.
\newblock Stratified transformer for 3d point cloud segmentation.
\newblock In \emph{Proceedings of the IEEE/CVF conference on computer vision and pattern recognition}, pages 8500--8509, 2022.

\bibitem[Lai et~al.(2023)Lai, Yuan, Chu, Chen, Hu, and Jia]{lai2023mask}
Xin Lai, Yuhui Yuan, Ruihang Chu, Yukang Chen, Han Hu, and Jiaya Jia.
\newblock Mask-attention-free transformer for 3d instance segmentation.
\newblock In \emph{Proceedings of the IEEE/CVF International Conference on Computer Vision}, pages 3693--3703, 2023.

\bibitem[Landrieu and Simonovsky(2018)]{Landrieu_Simonovsky_2018}
Loic Landrieu and Martin Simonovsky.
\newblock Large-scale point cloud semantic segmentation with superpoint graphs.
\newblock In \emph{Proceedings of the IEEE conference on computer vision and pattern recognition}, pages 4558--4567, 2018.

\bibitem[Lee et~al.(2023)Lee, Lee, Nam, Yu, Do, and Taghavi]{lee2023weakly}
Jungbeom Lee, Sungjin Lee, Jinseok Nam, Seunghak Yu, Jaeyoung Do, and Tara Taghavi.
\newblock Weakly supervised referring image segmentation with intra-chunk and inter-chunk consistency.
\newblock In \emph{Proceedings of the IEEE/CVF International Conference on Computer Vision}, pages 21870--21881, 2023.

\bibitem[Li et~al.(2023{\natexlab{a}})Li, Sun, Xiao, Lim, and Zhao]{li2023fully}
Hui Li, Mingjie Sun, Jimin Xiao, Eng~Gee Lim, and Yao Zhao.
\newblock Fully and weakly supervised referring expression segmentation with end-to-end learning.
\newblock \emph{IEEE Transactions on Circuits and Systems for Video Technology}, 2023{\natexlab{a}}.

\bibitem[Li et~al.(2022{\natexlab{a}})Li, He, Wei, Qian, Zhu, Xie, Zhuang, Tian, and Tang]{li2022fine}
Juncheng Li, Xin He, Longhui Wei, Long Qian, Linchao Zhu, Lingxi Xie, Yueting Zhuang, Qi~Tian, and Siliang Tang.
\newblock Fine-grained semantically aligned vision-language pre-training.
\newblock \emph{Advances in neural information processing systems}, 35:\penalty0 7290--7303, 2022{\natexlab{a}}.

\bibitem[Li et~al.(2022{\natexlab{b}})Li, Li, Xiong, and Hoi]{li2022blip}
Junnan Li, Dongxu Li, Caiming Xiong, and Steven Hoi.
\newblock Blip: Bootstrapping language-image pre-training for unified vision-language understanding and generation.
\newblock In \emph{International conference on machine learning}, pages 12888--12900. PMLR, 2022{\natexlab{b}}.

\bibitem[Li et~al.(2023{\natexlab{b}})Li, Li, Savarese, and Hoi]{li2023blip}
Junnan Li, Dongxu Li, Silvio Savarese, and Steven Hoi.
\newblock Blip-2: Bootstrapping language-image pre-training with frozen image encoders and large language models.
\newblock In \emph{International conference on machine learning}, pages 19730--19742. PMLR, 2023{\natexlab{b}}.

\bibitem[Li et~al.(2022{\natexlab{c}})Li, Tian, Shi, Chen, Zhao, Zhou, and Zhang]{li2022toist}
Pengfei Li, Beiwen Tian, Yongliang Shi, Xiaoxue Chen, Hao Zhao, Guyue Zhou, and Ya-Qin Zhang.
\newblock Toist: Task oriented instance segmentation transformer with noun-pronoun distillation.
\newblock \emph{Advances in Neural Information Processing Systems}, 35:\penalty0 17597--17611, 2022{\natexlab{c}}.

\bibitem[Li et~al.(2023{\natexlab{c}})Li, Chen, Zhao, Gong, Zhou, Rossano, and Zhu]{li2023understanding}
Yang Li, Xiaoxue Chen, Hao Zhao, Jiangtao Gong, Guyue Zhou, Federico Rossano, and Yixin Zhu.
\newblock Understanding embodied reference with touch-line transformer.
\newblock In \emph{ICLR}, 2023{\natexlab{c}}.

\bibitem[Li et~al.(2023{\natexlab{d}})Li, Wang, Xiao, Ji, and Chua]{li2023transformer}
Yicong Li, Xiang Wang, Junbin Xiao, Wei Ji, and Tat-Seng Chua.
\newblock Transformer-empowered invariant grounding for video question answering.
\newblock \emph{IEEE Transactions on Pattern Analysis and Machine Intelligence}, 2023{\natexlab{d}}.

\bibitem[Li et~al.(2024)Li, Zhao, Xiao, Feng, Wang, and Chua]{li2024laso}
Yicong Li, Na~Zhao, Junbin Xiao, Chun Feng, Xiang Wang, and Tat-seng Chua.
\newblock Laso: Language-guided affordance segmentation on 3d object.
\newblock In \emph{Proceedings of the IEEE/CVF Conference on Computer Vision and Pattern Recognition}, pages 14251--14260, 2024.

\bibitem[Liang et~al.(2021)Liang, Li, Xu, Tan, and Jia]{liang2021instance}
Zhihao Liang, Zhihao Li, Songcen Xu, Mingkui Tan, and Kui Jia.
\newblock Instance segmentation in 3d scenes using semantic superpoint tree networks.
\newblock In \emph{Proceedings of the IEEE/CVF International Conference on Computer Vision}, pages 2783--2792, 2021.

\bibitem[Lin et~al.(2023)Lin, Luo, Zheng, Li, Chao, Jin, Luo, Wang, Wang, and Cao]{lin2023unified}
Haojia Lin, Yongdong Luo, Xiawu Zheng, Lijiang Li, Fei Chao, Taisong Jin, Donghao Luo, Chengjie Wang, Yan Wang, and Liujuan Cao.
\newblock A unified framework for 3d point cloud visual grounding.
\newblock \emph{arXiv preprint arXiv:2308.11887}, 2023.

\bibitem[Liu et~al.(2023)Liu, Liu, Kong, Xu, Zhang, Yin, Hancke, and Lau]{liu2023referring}
Fang Liu, Yuhao Liu, Yuqiu Kong, Ke~Xu, Lihe Zhang, Baocai Yin, Gerhard Hancke, and Rynson Lau.
\newblock Referring image segmentation using text supervision.
\newblock In \emph{Proceedings of the IEEE/CVF International Conference on Computer Vision}, pages 22124--22134, 2023.

\bibitem[Liu et~al.(2019)Liu, Ott, Goyal, Du, Joshi, Chen, Levy, Lewis, Zettlemoyer, and Stoyanov]{liu2019roberta}
Yinhan Liu, Myle Ott, Naman Goyal, Jingfei Du, Mandar Joshi, Danqi Chen, Omer Levy, Mike Lewis, Luke Zettlemoyer, and Veselin Stoyanov.
\newblock Roberta: A robustly optimized bert pretraining approach.
\newblock \emph{arXiv preprint arXiv:1907.11692}, 2019.

\bibitem[Lu et~al.(2024)Lu, Pei, Wang, Li, Yang, Lei, and Shen]{lu2024scaneru}
Ziyang Lu, Yunqiang Pei, Guoqing Wang, Peiwei Li, Yang Yang, Yinjie Lei, and Heng~Tao Shen.
\newblock Scaneru: Interactive 3d visual grounding based on embodied reference understanding.
\newblock In \emph{Proceedings of the AAAI Conference on Artificial Intelligence}, volume~38, pages 3936--3944, 2024.

\bibitem[Luo et~al.(2022)Luo, Fu, Kong, Gao, Ren, Shen, Xia, and Liu]{Luo_Fu_Kong_Gao_Ren_Shen_Xia_Liu_2022}
Junyu Luo, Jiahui Fu, Xianghao Kong, Chen Gao, Haibing Ren, Hao Shen, Huaxia Xia, and Si~Liu.
\newblock 3d-sps: Single-stage 3d visual grounding via referred point progressive selection.
\newblock In \emph{Proceedings of the IEEE/CVF Conference on Computer Vision and Pattern Recognition}, pages 16454--16463, 2022.

\bibitem[Manning et~al.(2014)Manning, Surdeanu, Bauer, Finkel, Bethard, and McClosky]{Manning_Surdeanu_Bauer_Finkel_Bethard_McClosky_2014}
Christopher~D Manning, Mihai Surdeanu, John Bauer, Jenny~Rose Finkel, Steven Bethard, and David McClosky.
\newblock The stanford corenlp natural language processing toolkit.
\newblock In \emph{Proceedings of 52nd annual meeting of the association for computational linguistics: system demonstrations}, pages 55--60, 2014.

\bibitem[Milletari et~al.(2016)Milletari, Navab, and Ahmadi]{Milletari_Navab_Ahmadi_2016}
Fausto Milletari, Nassir Navab, and Seyed-Ahmad Ahmadi.
\newblock V-net: Fully convolutional neural networks for volumetric medical image segmentation.
\newblock In \emph{2016 fourth international conference on 3D vision (3DV)}, pages 565--571. Ieee, 2016.

\bibitem[Mithun et~al.(2019)Mithun, Paul, and Roy-Chowdhury]{mithun2019weakly}
Niluthpol~Chowdhury Mithun, Sujoy Paul, and Amit~K Roy-Chowdhury.
\newblock Weakly supervised video moment retrieval from text queries.
\newblock In \emph{Proceedings of the IEEE/CVF Conference on Computer Vision and Pattern Recognition}, pages 11592--11601, 2019.

\bibitem[Mo and Liu(2024)]{mo2024bridging}
Wentao Mo and Yang Liu.
\newblock Bridging the gap between 2d and 3d visual question answering: A fusion approach for 3d vqa.
\newblock \emph{arXiv preprint arXiv:2402.15933}, 2024.

\bibitem[Qi et~al.(2019)Qi, Litany, He, and Guibas]{Qi_Litany_He_Guibas_2019}
Charles~R Qi, Or~Litany, Kaiming He, and Leonidas~J Guibas.
\newblock Deep hough voting for 3d object detection in point clouds.
\newblock In \emph{proceedings of the IEEE/CVF International Conference on Computer Vision}, pages 9277--9286, 2019.

\bibitem[Qi et~al.(2017)Qi, Yi, Su, and Guibas]{qi2017pointnet++}
Charles~Ruizhongtai Qi, Li~Yi, Hao Su, and Leonidas~J Guibas.
\newblock Pointnet++: Deep hierarchical feature learning on point sets in a metric space.
\newblock \emph{Advances in neural information processing systems}, 30, 2017.

\bibitem[Qian et~al.(2024)Qian, Ma, Ji, and Sun]{qian2024x}
Zhipeng Qian, Yiwei Ma, Jiayi Ji, and Xiaoshuai Sun.
\newblock X-refseg3d: Enhancing referring 3d instance segmentation via structured cross-modal graph neural networks.
\newblock In \emph{Proceedings of the AAAI Conference on Artificial Intelligence}, volume~38, pages 4551--4559, 2024.

\bibitem[Radford et~al.(2021)Radford, Kim, Hallacy, Ramesh, Goh, Agarwal, Sastry, Askell, Mishkin, Clark, et~al.]{radford2021learning}
Alec Radford, Jong~Wook Kim, Chris Hallacy, Aditya Ramesh, Gabriel Goh, Sandhini Agarwal, Girish Sastry, Amanda Askell, Pamela Mishkin, Jack Clark, et~al.
\newblock Learning transferable visual models from natural language supervision.
\newblock In \emph{International conference on machine learning}, pages 8748--8763. PMLR, 2021.

\bibitem[Redmon et~al.(2016)Redmon, Divvala, Girshick, and Farhadi]{redmon2016you}
Joseph Redmon, Santosh Divvala, Ross Girshick, and Ali Farhadi.
\newblock You only look once: Unified, real-time object detection.
\newblock In \emph{Proceedings of the IEEE conference on computer vision and pattern recognition}, pages 779--788, 2016.

\bibitem[Schult et~al.(2022)Schult, Engelmann, Hermans, Litany, Tang, and Leibe]{schult2022mask3d}
Jonas Schult, Francis Engelmann, Alexander Hermans, Or~Litany, Siyu Tang, and Bastian Leibe.
\newblock Mask3d for 3d semantic instance segmentation.
\newblock \emph{arXiv preprint arXiv:2210.03105}, 2022.

\bibitem[Song et~al.(2020)Song, Tan, Qin, Lu, and Liu]{song2020mpnet}
Kaitao Song, Xu~Tan, Tao Qin, Jianfeng Lu, and Tie-Yan Liu.
\newblock Mpnet: Masked and permuted pre-training for language understanding.
\newblock \emph{Advances in Neural Information Processing Systems}, 33:\penalty0 16857--16867, 2020.

\bibitem[Subramanian et~al.(2022)Subramanian, Merrill, Darrell, Gardner, Singh, and Rohrbach]{subramanian-etal-2022-reclip}
Sanjay Subramanian, William Merrill, Trevor Darrell, Matt Gardner, Sameer Singh, and Anna Rohrbach.
\newblock {R}e{CLIP}: A strong zero-shot baseline for referring expression comprehension.
\newblock In Smaranda Muresan, Preslav Nakov, and Aline Villavicencio, editors, \emph{Proceedings of the 60th Annual Meeting of the Association for Computational Linguistics (Volume 1: Long Papers)}, pages 5198--5215, Dublin, Ireland, May 2022. Association for Computational Linguistics.
\newblock \doi{10.18653/v1/2022.acl-long.357}.
\newblock URL \url{https://aclanthology.org/2022.acl-long.357}.

\bibitem[Sun et~al.(2023)Sun, Qing, Tan, and Xu]{Sun_Qing_Tan_Xu_2022}
Jiahao Sun, Chunmei Qing, Junpeng Tan, and Xiangmin Xu.
\newblock Superpoint transformer for 3d scene instance segmentation.
\newblock In \emph{Proceedings of the AAAI Conference on Artificial Intelligence}, volume~37, pages 2393--2401, 2023.

\bibitem[Suo et~al.(2023)Suo, Zhu, and Yang]{suo-etal-2023-text}
Yucheng Suo, Linchao Zhu, and Yi~Yang.
\newblock Text augmented spatial aware zero-shot referring image segmentation.
\newblock In Houda Bouamor, Juan Pino, and Kalika Bali, editors, \emph{Findings of the Association for Computational Linguistics: EMNLP 2023}, pages 1032--1043, Singapore, December 2023. Association for Computational Linguistics.
\newblock \doi{10.18653/v1/2023.findings-emnlp.73}.
\newblock URL \url{https://aclanthology.org/2023.findings-emnlp.73}.

\bibitem[Trott et~al.(2017)Trott, Xiong, and Socher]{trott2017interpretable}
Alexander Trott, Caiming Xiong, and Richard Socher.
\newblock Interpretable counting for visual question answering.
\newblock \emph{arXiv preprint arXiv:1712.08697}, 2017.

\bibitem[Vaswani et~al.(2017)Vaswani, Shazeer, Parmar, Uszkoreit, Jones, Gomez, Kaiser, and Polosukhin]{vaswani2017attention}
Ashish Vaswani, Noam Shazeer, Niki Parmar, Jakob Uszkoreit, Llion Jones, Aidan~N Gomez, {\L}ukasz Kaiser, and Illia Polosukhin.
\newblock Attention is all you need.
\newblock \emph{Advances in neural information processing systems}, 30, 2017.

\bibitem[Wang et~al.(2023)Wang, Huang, Zhao, Zhang, and Zhao]{wang2023chat}
Zehan Wang, Haifeng Huang, Yang Zhao, Ziang Zhang, and Zhou Zhao.
\newblock Chat-3d: Data-efficiently tuning large language model for universal dialogue of 3d scenes.
\newblock \emph{arXiv preprint arXiv:2308.08769}, 2023.

\bibitem[Wu et~al.(2023{\natexlab{a}})Wu, Ma, Chen, Wang, Luo, Ji, and Sun]{wu20233d}
Changli Wu, Yiwei Ma, Qi~Chen, Haowei Wang, Gen Luo, Jiayi Ji, and Xiaoshuai Sun.
\newblock 3d-stmn: Dependency-driven superpoint-text matching network for end-to-end 3d referring expression segmentation.
\newblock \emph{arXiv preprint arXiv:2308.16632}, 2023{\natexlab{a}}.

\bibitem[Wu et~al.(2024{\natexlab{a}})Wu, Liu, Ji, Ma, Wang, Luo, Ding, Sun, and Ji]{wu20243dgresgeneralized3dreferring}
Changli Wu, Yihang Liu, Jiayi Ji, Yiwei Ma, Haowei Wang, Gen Luo, Henghui Ding, Xiaoshuai Sun, and Rongrong Ji.
\newblock 3d-gres: Generalized 3d referring expression segmentation, 2024{\natexlab{a}}.
\newblock URL \url{https://arxiv.org/abs/2407.20664}.

\bibitem[Wu et~al.(2021)Wu, Peng, Chen, Fu, and Chao]{wu2021rethinking}
Kan Wu, Houwen Peng, Minghao Chen, Jianlong Fu, and Hongyang Chao.
\newblock Rethinking and improving relative position encoding for vision transformer.
\newblock In \emph{Proceedings of the IEEE/CVF International Conference on Computer Vision}, pages 10033--10041, 2021.

\bibitem[Wu et~al.(2024{\natexlab{b}})Wu, Fei, Li, Ji, Zhang, Chua, and Yan]{wu2024towards}
Shengqiong Wu, Hao Fei, Xiangtai Li, Jiayi Ji, Hanwang Zhang, Tat-Seng Chua, and Shuicheng Yan.
\newblock Towards semantic equivalence of tokenization in multimodal llm.
\newblock \emph{arXiv preprint arXiv:2406.05127}, 2024{\natexlab{b}}.

\bibitem[Wu et~al.(2024{\natexlab{c}})Wu, Fei, Qu, Ji, and Chua]{wu24next}
Shengqiong Wu, Hao Fei, Leigang Qu, Wei Ji, and Tat-Seng Chua.
\newblock Next-gpt: Any-to-any multimodal llm.
\newblock In \emph{Proceedings of the International Conference on Machine Learning}, 2024{\natexlab{c}}.

\bibitem[Wu et~al.(2023{\natexlab{b}})Wu, Cheng, Zhang, Cheng, and Zhang]{Wu_Cheng_Zhang_Cheng_Zhang_2022}
Yanmin Wu, Xinhua Cheng, Renrui Zhang, Zesen Cheng, and Jian Zhang.
\newblock Eda: Explicit text-decoupling and dense alignment for 3d visual grounding.
\newblock In \emph{Proceedings of the IEEE/CVF Conference on Computer Vision and Pattern Recognition}, pages 19231--19242, 2023{\natexlab{b}}.

\bibitem[Xu et~al.(2024)Xu, Shi, Tu, Zhou, Liang, and Bai]{xu2024unified}
Wei Xu, Chunsheng Shi, Sifan Tu, Xin Zhou, Dingkang Liang, and Xiang Bai.
\newblock A unified framework for 3d scene understanding.
\newblock \emph{arXiv preprint arXiv:2407.03263}, 2024.

\bibitem[Yang et~al.(2021)Yang, Zhang, Wang, and Luo]{Yang_Zhang_Wang_Luo_2021}
Zhengyuan Yang, Songyang Zhang, Liwei Wang, and Jiebo Luo.
\newblock Sat: 2d semantics assisted training for 3d visual grounding.
\newblock In \emph{Proceedings of the IEEE/CVF International Conference on Computer Vision}, pages 1856--1866, 2021.

\bibitem[Yuan et~al.(2021)Yuan, Yan, Liao, Zhang, Wang, Li, and Cui]{Yuan_Yan_Liao_Zhang_Wang_Li_Cui_2021}
Zhihao Yuan, Xu~Yan, Yinghong Liao, Ruimao Zhang, Sheng Wang, Zhen Li, and Shuguang Cui.
\newblock Instancerefer: Cooperative holistic understanding for visual grounding on point clouds through instance multi-level contextual referring.
\newblock In \emph{Proceedings of the IEEE/CVF International Conference on Computer Vision}, pages 1791--1800, 2021.

\bibitem[Zhang et~al.(2023)Zhang, Yannakoudakis, Zhen, and Shutova]{zhang-etal-2023-ck}
Zhi Zhang, Helen Yannakoudakis, Xiantong Zhen, and Ekaterina Shutova.
\newblock {CK}-transformer: Commonsense knowledge enhanced transformers for referring expression comprehension.
\newblock In Andreas Vlachos and Isabelle Augenstein, editors, \emph{Findings of the Association for Computational Linguistics: EACL 2023}, pages 2586--2596, Dubrovnik, Croatia, May 2023. Association for Computational Linguistics.
\newblock \doi{10.18653/v1/2023.findings-eacl.196}.
\newblock URL \url{https://aclanthology.org/2023.findings-eacl.196}.

\bibitem[Zhao et~al.(2021)Zhao, Cai, Sheng, and Xu]{Zhao_Cai_Sheng_Xu_2021}
Lichen Zhao, Daigang Cai, Lu~Sheng, and Dong Xu.
\newblock 3dvg-transformer: Relation modeling for visual grounding on point clouds.
\newblock In \emph{Proceedings of the IEEE/CVF International Conference on Computer Vision}, pages 2928--2937, 2021.

\bibitem[Zheng et~al.(2023)Zheng, Huang, Zhao, Zhong, and Wang]{zheng2023towards}
Duo Zheng, Shijia Huang, Lin Zhao, Yiwu Zhong, and Liwei Wang.
\newblock Towards learning a generalist model for embodied navigation.
\newblock \emph{arXiv preprint arXiv:2312.02010}, 2023.

\bibitem[Zhu et~al.(2023)Zhu, Ma, Chen, Deng, Huang, and Li]{zhu20233d}
Ziyu Zhu, Xiaojian Ma, Yixin Chen, Zhidong Deng, Siyuan Huang, and Qing Li.
\newblock 3d-vista: Pre-trained transformer for 3d vision and text alignment.
\newblock In \emph{Proceedings of the IEEE/CVF International Conference on Computer Vision}, pages 2911--2921, 2023.

\end{thebibliography}
}

\clearpage
\appendix
{\LARGE \textbf{Appendix}}

\section{The Critical Role of Spatial Information in 3D-RES Tasks}
Our analysis underscores the pivotal role spatial relations play in 3D-RES tasks. We assessed the ScanRefer dataset's referring expressions, classifying examples into two categories: those with spatial relation terms (e.g., ``left'', ``right'', ``next'', ``bottom'' and ``side") as spatially related, and those without as spatially unrelated. Our findings revealed that spatially related samples form about 92\% of the dataset, highlighting the prevalence of spatial descriptors. Additionally, the Sr3D dataset consists entirely of spatially related descriptions, and in the Nr3D dataset, a significant 90.5\% of entries utilize spatial prepositions~\cite{Achlioptas_Abdelreheem_Xia_Elhoseiny_Guibas_2020}. This evidence demonstrates the necessity of spatial descriptions in accurately identifying objects within a scene through natural language, emphasizing the essential need for effective spatial relation modeling in 3D-RES tasks.

\begin{wrapfigure}{r}{0.4\textwidth}
    \centering
    \includegraphics[width=0.4\columnwidth]{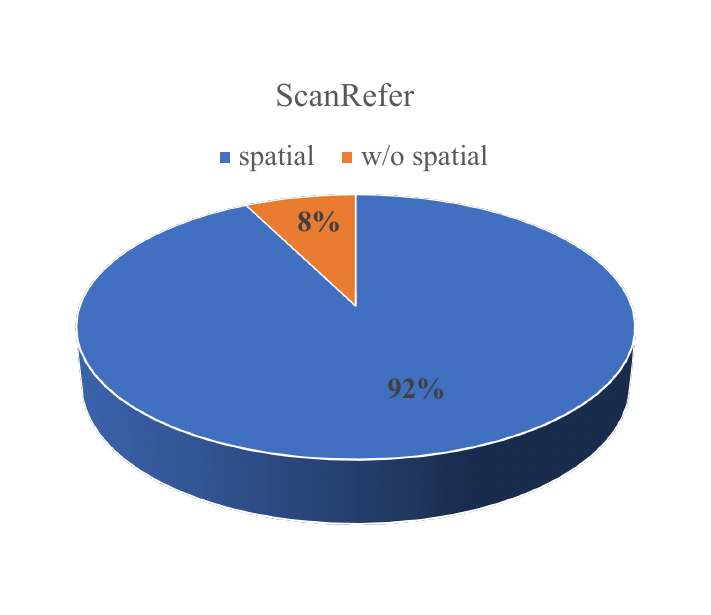}
    \caption{Statistics of samples in the ScanRefer dataset based on the presence of spatial relation descriptions, where ``spatial'' represents samples with spatially related descriptions, while ``w/o spatial'' denotes spatially unrelated samples.}
    \label{fig:sp_sta}
\end{wrapfigure}

\section{3D-RES on ReferIt3D Dataset}
We extended the 3D-RES task on the ReferIt3D dataset~\cite{Achlioptas_Abdelreheem_Xia_Elhoseiny_Guibas_2020} (which is also in English) by integrating instance masks from ScanNet 
and conducting relevant experiments, as shown in Tab.~\ref{tab:referit3d}. In contrast to the original setup of ReferIt3D, we refrained from using ground truth bounding boxes or masks as input during our experiments, which significantly increased the level of difficulty. Nonetheless, our model achieved remarkable Acc@50 gains of \textbf{5.3} points for Sr3D and \textbf{2.9} points for Nr3D, accompanied by mIoU gains of 5.2 points for Sr3D and 1.0 points for Nr3D.

It is worth highlighting that our results demonstrate exceptional performance in terms of Acc@50 and mIoU. This can be attributed to the incorporation of spatial information, which enhances the accuracy of segmentation results and addresses the challenges of over-segmentation and under-segmentation encountered in previous approaches.

\section{More Ablation Studies}
\subsection{Number of Multiple Rounds}
We investigated the impact of varying the number of TLM rounds in our model. Analyzing rows two to five in Tab.~\ref{tab:num_layers} reveals a consistent pattern: performance improves with more rounds, reaches its peak at six, and then slightly declines. Fewer layers result in insufficient capacity, while an excessive number of layers increases the risk of overfitting. Therefore, selecting six layers strikes a balance that yields the best model performance.

In addition, we conducted ablation experiments to remove the iterative position refinement process at each layer. The results, shown in the first row of Tab.~\ref{tab:num_layers}, clearly demonstrate the effectiveness of iterative refinement, leading to a significant improvement.

\begin{table*}
    \caption{Results of 3D-RES tasks on ReferIt3D.}
    \centering
    \footnotesize
    \centering
    \resizebox{1\textwidth}{!}{
    \begin{tabular}{c|ccc|ccc|ccc|ccc|ccc|c}
    \toprule
     & \multicolumn{3}{c|}{easy} & \multicolumn{3}{c|}{hard} & \multicolumn{3}{c|}{View Dep} & \multicolumn{3}{c|}{View Indep} & \multicolumn{3}{c|}{\textbf{Overall}} & {Inference} \\
     \multirow{-2}{*}{Method} & 0.25 & 0.5 & mIoU & 0.25 & 0.5 & mIoU & 0.25 & 0.5 & mIoU & 0.25 & 0.5 & mIoU & \textbf{0.25} & \textbf{0.5} & \textbf{mIoU} & Time\\
    \hline\hline 
    \multicolumn{17}{c}{\textbf{Nr3D}}\\
    \midrule
    TGNN & 29.2 & 22.3 & 21.0
    & 22.5 & 19.6 & 17.4
    & 22.2 & 18.1 & 17.0
    & 27.6 & 22.4 & 20.3
    & 25.7 & 20.9 & 19.1 & 26599ms\\  
    3D-STMN  & \textbf{47.9} & 31.9 & 32.6
    & \textbf{35.4} & 20.0 & 23.0
    & \textbf{37.7} & 21.1 & 24.3
    & \textbf{43.5} & 28.4 & 29.4
    & \textbf{41.5} & {25.8} & {27.6}   & {276ms} \\ 
    RG-SAN & 45.8 & \textbf{34.5} & \textbf{33.5}
    & {34.5} & \textbf{23.3} & \textbf{24.0}
    & {37.3} & \textbf{26.3} & \textbf{26.3}
    & 41.5 & \textbf{30.1} & \textbf{29.8}
    & 40.1 & \textbf{28.7} & \textbf{28.6} & 289ms\\
    \hline\hline 
    \multicolumn{17}{c}{\textbf{Sr3D}}\\
    \midrule
    TGNN & 28.2 & 23.0 & 20.9
    & 29.1 & 25.8 & 21.9
    & 23.8 & 21.3 & 18.2
    & 28.6 & 23.9 & 21.3
    & 27.5 & 22.9 & 20.2 & 26674ms\\  
    3D-STMN & 49.4 & 38.2 & 36.3 
    & 41.9 & 31.0 & 30.1 
    & \textbf{45.5} & 33.5 & 31.9
    & 47.2 & 36.2 & 34.6
    & {47.2} & {36.1} & {34.4}  & {281ms} \\
    RG-SAN & \textbf{55.8} & \textbf{43.6} & \textbf{41.5}
    & \textbf{46.7} & \textbf{36.3} & \textbf{35.2}
    & 42.6 & \textbf{34.7} & \textbf{32.5}
    & \textbf{53.5} & \textbf{41.7} & \textbf{39.9}
    & \textbf{53.1} & \textbf{41.4} & \textbf{39.6} & 293ms \\
    \bottomrule
    \end{tabular}
    }
    \label{tab:referit3d}
\end{table*}

\begin{table}
\caption{Number of multiple rounds.}
\centering
\begin{tabular}{c|c|ccc|ccc}
\toprule
& Number of & \multicolumn{3}{c|}{Multiple} 
& \multicolumn{3}{c}{\textbf{Overall}}
\\
\multirow{-2}{*}{Iter. Refine} & {Rounds} 
& 0.25 & 0.5 & mIoU 
& \textbf{0.25} & \textbf{0.5} & \textbf{mIoU} \\
\midrule

& 6 & 55.2 & 33.0 & 36.3 & 61.8 & 42.9 & 43.7 \\

\midrule

\checkmark & 1 & \textbf{56.3} & 30.3 & 35.6 & \textbf{62.8} & 40.6 & 43.1 \\  

\checkmark & 3 & 55.2 & 34.5 & 37.1 & 61.7 & 44.1 & 44.3 \\

\checkmark & 6 & {55.0} & {35.4} & \textbf{37.4}  & {61.7} & {44.9} & \textbf{44.6} \\

\checkmark & 9 & 53.1 & \textbf{35.8} & 37.0 & 60.0 & \textbf{45.2} & 44.2 \\

\bottomrule
\end{tabular}
\label{tab:num_layers}
\end{table}

\begin{table}
\centering
\caption{Ablation study comparing text encoders.}
\begin{tabular}{c|ccc|ccc}
\toprule
& \multicolumn{3}{c|}{Multiple} 
& \multicolumn{3}{c}{\textbf{Overall}}
\\
\multirow{-2}{*}{Text Encoder} 
& 0.25 & 0.5 & mIoU 
& \textbf{0.25} & \textbf{0.5} & \textbf{mIoU} \\
\midrule

BERT~\cite{Devlin_Chang_Lee_Toutanova_2019} & 53.2 & 34.8 & 36.4 & 60.0 & 44.1 & 43.7 \\  

RoBERTa~\cite{liu2019roberta} & 53.1 & 34.8 & 36.7 & 60.0 & 44.0 & 44.0 \\

CLIP~\cite{radford2021learning} & 52.3 & 33.4 & 35.0 & 58.2 & 42.1 & 42.5 \\

MPNet~\cite{song2020mpnet} & \textbf{55.0} & \textbf{35.4} & \textbf{37.4}  & \textbf{61.7} & \textbf{44.9} & \textbf{44.6} \\

\bottomrule
\end{tabular}
\label{tab:text_encoder}
\end{table}

\begin{table}
\centering
\caption{Ablation study comparing visual backbones.}
 \begin{tabular}{c|ccc|ccc}
\toprule
& \multicolumn{3}{c|}{Multiple} 
& \multicolumn{3}{c}{\textbf{Overall}}
\\
\multirow{-2}{*}{Text Encoder} 
& 0.25 & 0.5 & mIoU 
& \textbf{0.25} & \textbf{0.5} & \textbf{mIoU} \\
\midrule

SSTNet~\cite{Devlin_Chang_Lee_Toutanova_2019} & 53.7 & 34.3 & 34.9 & 59.4 & 42.5 & 43.2 \\  

PointNet++~\cite{liu2019roberta} & 54.1 & 34.6 & 36.1 & 60.3 & 44.2 & 44.0 \\

SPFormer~\cite{radford2021learning} & \textbf{55.0} & \textbf{35.4} & \textbf{37.4}  & \textbf{61.7} & \textbf{44.9} & \textbf{44.6} \\

\bottomrule
\end{tabular}
\label{tab:visual_encoder}
\end{table}

\subsection{The Textual Backbone}

In Tab.~\ref{tab:text_encoder}, we compare the effects of commonly used natural language encoders. It can be observed that our method demonstrates robustness with respect to the selection of the NLP backbone. And we achieve the best performance using MPNet~\cite{song2020mpnet}. The underperformance of CLIP~\cite{radford2021learning}is understandable, considering its optimization over a large dataset of text-image pairs. While CLIP excels at extracting representations at the sentence level, it encounters difficulties in comprehending intricate spatial relationships within sentences. This limitation hampers its performance in situations involving multiple objects, resulting in relatively poorer results.

\subsection{The Visual Backbone}
We explored alternative visual backbones, including the PointNet++~\cite{qi2017pointnet++} pretrained by the classic work 3D-VisTA~\cite{zhu20233d} and another superpoint-based backbone, SSTNet~\cite{liang2021instance}, as detailed in Tab.~\ref{tab:visual_encoder}. Our findings indicate that the performance with PointNet++, SSTNet and our employed SPFormer are quite comparable, demonstrating the adaptability and effectiveness of our proposed modules across different backbone architectures.

\section{More Qualitative Analysis}
\begin{figure*}
    \centering
    \includegraphics[width=1.0\columnwidth]{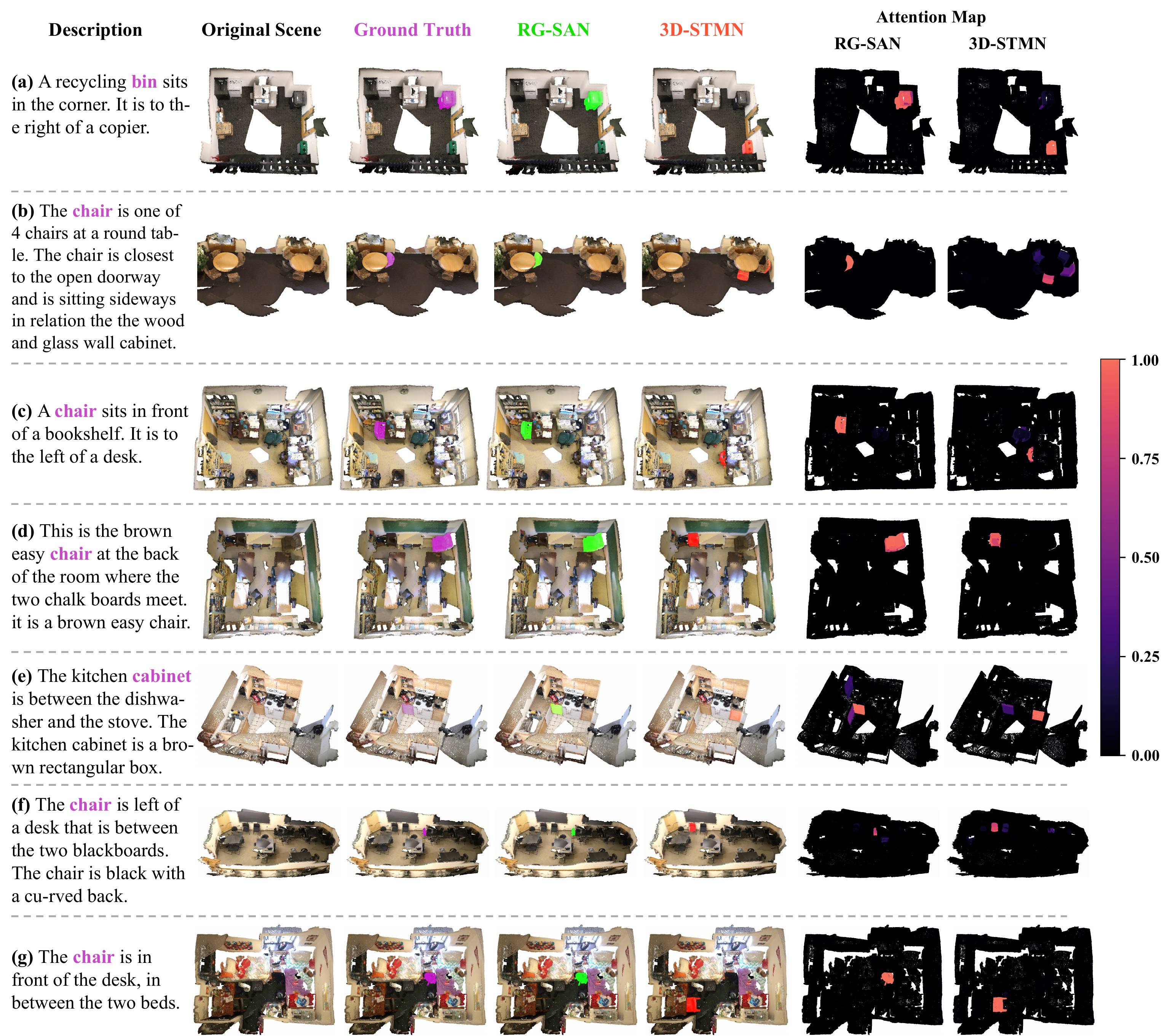}
    \caption{Qualitative comparison between the proposed RG-SAN and 3D-STMN. \textbf{Zoom in for best view.}}
    \label{fig:5_nips}
\end{figure*}

\begin{figure*}
    \centering
    \includegraphics[width=1.0\columnwidth]{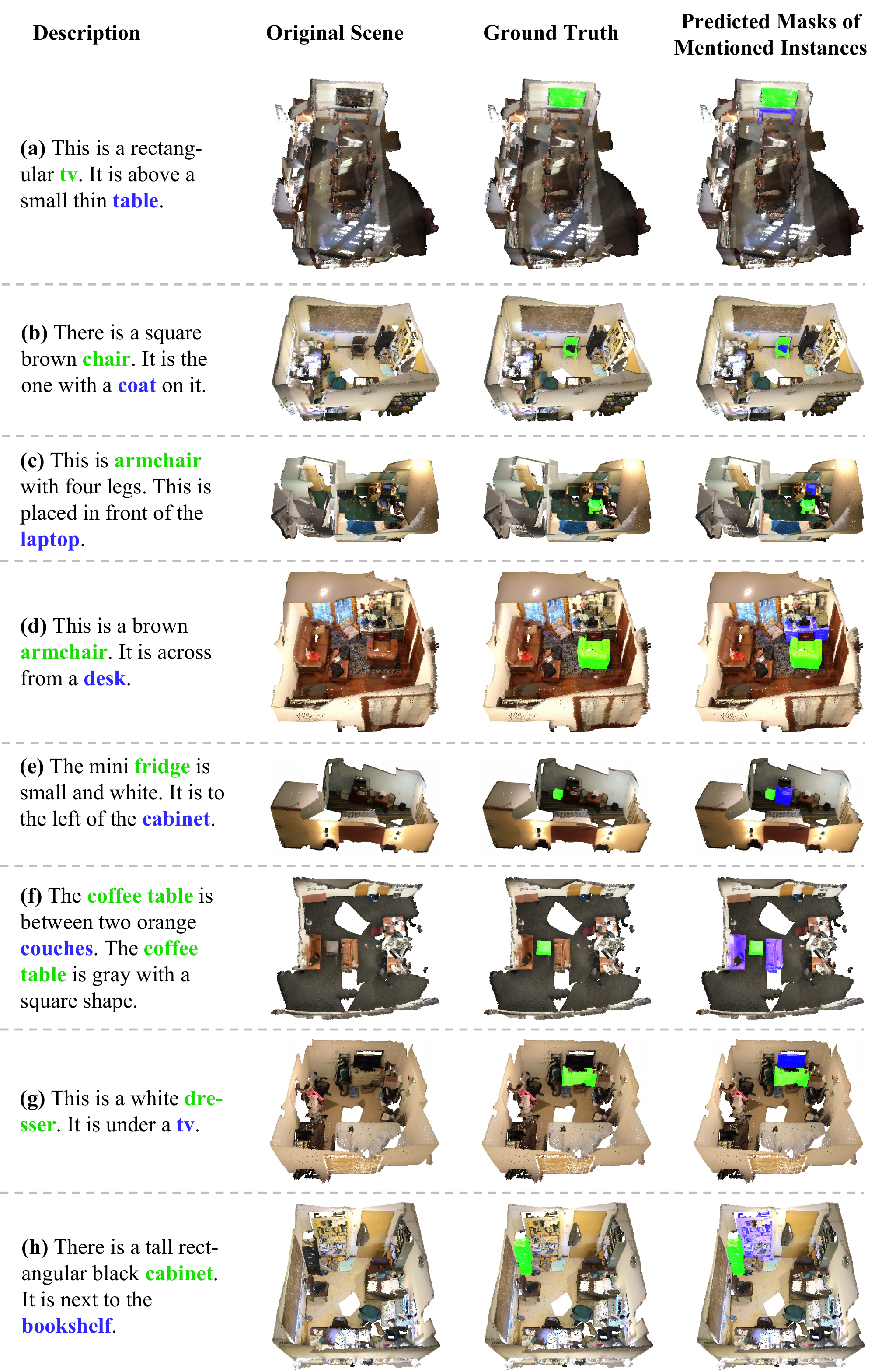}
    \caption{The visualization of the predicted masks of mentioned instances of our RG-SAN. \textbf{Zoom in for best view.}}
    \label{fig:6}
\end{figure*}

More qualitative comparison results are illustrated in Fig.~\ref{fig:5_nips} and Fig.~\ref{fig:6}, demonstrating the remarkable discriminative ability of our RG-SAN compared to 3D-STMN. Fig.~\ref{fig:5_nips} showcases RG-SAN's superior performance in accurately localizing target objects, especially in challenging scenarios that require understanding complex positional relationships described in the text. For instance, Fig.~\ref{fig:5_nips}-\textbf{(b)} illustrates a scenario with numerous distractors and a complex textual description, where 3D-STMN fails, causing over-segmentation. In contrast, RG-SAN accurately discerns and localizes the target object amidst distractions, achieving higher-quality segmentation. It is important to highlight that when faced with descriptive text that involves spatial relationship reasoning among multiple instances mentioned, as seen in all cases in Fig.~\ref{fig:5_nips}, our RG-SAN demonstrates the capability to precisely locate and identify the target object. In contrast, 3D-STMN\cite{wu20233d} lacks comparable complex reasoning abilities in such scenarios.

In Fig.~\ref{fig:6}, we visualized the predicted masks of the mentioned instances of our RG-SAN. As can be seen in Fig.~\ref{fig:6}~\textbf{(b)}, even if the ``coat'' category is not present in the training labels, our RG-SAN is still able to accurately identify the mentioned ``coat'' in the point cloud scene. This is because we align the word features from the textual modality and the point cloud features from the visual modality in a fine-grained manner through weak supervision. This alignment brings them into the same feature space, enabling the model to have strong generalization capabilities for unknown semantic categories. This paves the way for future research in weak supervision and open vocabulary.

Furthermore, as seen in Fig.~\ref{fig:6}~\textbf{(f)}, our RG-SAN is even able to accurately recognize the plural form of the entity noun ``couches'' mentioned in the descriptive text, while successfully identifying the target object. This capability enables the model to have a more precise and efficient understanding of spatial relationships associated with multiple auxiliary objects, such as ``between'' and ``among'', showcasing the powerful spatial relationship modeling ability of our RG-SAN.

\section{Analysis of Target Word Positioning Capability in LLMs}
We attempt to utilize LLMs, specifically LLAMA 2 70B as an example, for target word positioning. To achieve this, we construct a command template for LLMs, which includes the input description token list and demonstration examples. Such a general template is designed as follows:
\begin{align}\small
\begin{split}
&\operatorname{``\ Given\ a\ word\ list,\ find\ the\ target\ word\ in\ the\ list:}\\
&\ \ \ \ \ \ \ \ \operatorname{[`the',`trash',`can',`is',`directly',`right',}\\
&\ \ \ \ \ \ \ \ \operatorname{`of',`the',`brown',`tables',`turned',}\\
&\ \ \ \ \ \ \ \ \operatorname{`sideways',`.'] => `can',\ 2}\\
&\ \ \ \ \ \ \ \ \operatorname{[`there',`is',`a',`dark',`brown',`wooden',}\\
&\ \ \ \ \ \ \ \ \operatorname{`and',`leather',`chair',`.',`placed',`in',`the',}\\
&\ \ \ \ \ \ \ \ \operatorname{`table',`of',`the',`kitchen',`.'] => `chair',\ 8}\\
&\ \ \ \ \ \ \ \ \operatorname{\textcolor{gray}{[LIST]}:\ =>''}
\end{split}
\label{template1}
\end{align}
where $\operatorname{\textcolor{gray}{[LIST]}}$ is replaced by the input description token list, and `` $\operatorname{=> `can',\ 2}$ '' denotes the target token in the first example is ``can'' whose index in the token list is 2.

For the ScanRefer dataset, LLAMA 2 70B produces approximately 80\% of target word positions that align with the results obtained from our RWS module. In the remaining portion, our RWS module demonstrates higher accuracy. This partially indicates that there is still room for improvement in LLAMA 2 70B's ability to identify target word positions. Conversely, our rule-based approach benefits from efficient utilization of explicit dependency relationships and exhibits certain advantages. Additionally, LLAMA 2 70B poses a significant computational burden. Taking this into consideration, our adopted RWS approach outperforms LLAMA 2 70B in terms of both accuracy and efficiency.

\section{Limitations and Broader Impact}
\label{limitations}
Despite the strong performance of RG-SAN, we identify several limitations that call for further improvement. A primary limitation is its difficulty in accurately localizing plural nouns. This issue arises from the method of using a single point for localization, which proves challenging for plural entities in certain contexts. In future work, we will explore using multiple points to delineate boundaries for more precise localization of plural nouns.

The second limitation of our study concerns the model's inadequate robustness towards damaged point cloud data. The occurrence of damaged or incomplete data within point cloud datasets presents a significant challenge, one that our current model is not sufficiently equipped to address. This lack of robustness can impair the model's ability to process such data accurately, leading to unreliable results in scenarios involving incomplete or corrupted point clouds. Future work will aim to enhance the model's resilience and capability in handling and compensating for data imperfections.

RG-SAN is expected to stimulate further development and application of multimodal 3D perception, especially in practical scenarios such as embodied intelligence and autonomous driving. However, when it comes to practical applications, particularly those involving safety and privacy, rigorous testing is required to ensure compliance with relevant laws and regulations.

\section{Ethics Statement and Licenses}
In our work, there are no human subjects and informed consent is not applicable. Additionally, we use publicly available text data from the ScanRefer Dataset (\url{https://daveredrum.github.io/ScanRefer}), which is licensed under a \textit{Creative Commons Attribution-NonCommercial-ShareAlike 3.0 Unported License} which allows us to use the dataset for non-commercial purposes. For point cloud data, we used the publicly available ScanNet Dataset (\url{https://github.com/ScanNet/ScanNet}), which is licensed under the \href{http://kaldir.vc.in.tum.de/scannet/ScanNet_TOS.pdf}{ScanNet Terms of Use}, and the code is released under the MIT license. Both the licenses of ScanNet allow us to use the dataset and code for non-commercial purposes. In the appendix, we use the ReferIt3D Dataset (\url{https://github.com/referit3d/referit3d}) for extra experiments, which is licensed under the MIT license which allows us to use the dataset for non-commercial purposes.

\clearpage

\newpage
\section*{NeurIPS Paper Checklist}

\begin{enumerate}

\item {\bf Claims}
    \item[] Question: Do the main claims made in the abstract and introduction accurately reflect the paper's contributions and scope?
    \item[] Answer: \answerYes{} 
    \item[] Justification: The abstract and introduction accurately summarize the paper's key contributions and scope.
    \item[] Guidelines:
    \begin{itemize}
        \item The answer NA means that the abstract and introduction do not include the claims made in the paper.
        \item The abstract and/or introduction should clearly state the claims made, including the contributions made in the paper and important assumptions and limitations. A No or NA answer to this question will not be perceived well by the reviewers. 
        \item The claims made should match theoretical and experimental results, and reflect how much the results can be expected to generalize to other settings. 
        \item It is fine to include aspirational goals as motivation as long as it is clear that these goals are not attained by the paper. 
    \end{itemize}

\item {\bf Limitations}
    \item[] Question: Does the paper discuss the limitations of the work performed by the authors?
    \item[] Answer: \answerYes{}{} 
    \item[] Justification: We discuss the limitations of the work in the Appendix.
    \item[] Guidelines:
    \begin{itemize}
        \item The answer NA means that the paper has no limitation while the answer No means that the paper has limitations, but those are not discussed in the paper. 
        \item The authors are encouraged to create a separate "Limitations" section in their paper.
        \item The paper should point out any strong assumptions and how robust the results are to violations of these assumptions (e.g., independence assumptions, noiseless settings, model well-specification, asymptotic approximations only holding locally). The authors should reflect on how these assumptions might be violated in practice and what the implications would be.
        \item The authors should reflect on the scope of the claims made, e.g., if the approach was only tested on a few datasets or with a few runs. In general, empirical results often depend on implicit assumptions, which should be articulated.
        \item The authors should reflect on the factors that influence the performance of the approach. For example, a facial recognition algorithm may perform poorly when image resolution is low or images are taken in low lighting. Or a speech-to-text system might not be used reliably to provide closed captions for online lectures because it fails to handle technical jargon.
        \item The authors should discuss the computational efficiency of the proposed algorithms and how they scale with dataset size.
        \item If applicable, the authors should discuss possible limitations of their approach to address problems of privacy and fairness.
        \item While the authors might fear that complete honesty about limitations might be used by reviewers as grounds for rejection, a worse outcome might be that reviewers discover limitations that aren't acknowledged in the paper. The authors should use their best judgment and recognize that individual actions in favor of transparency play an important role in developing norms that preserve the integrity of the community. Reviewers will be specifically instructed to not penalize honesty concerning limitations.
    \end{itemize}

\item {\bf Theory Assumptions and Proofs}
    \item[] Question: For each theoretical result, does the paper provide the full set of assumptions and a complete (and correct) proof?
    \item[] Answer: \answerNA{} 
    \item[] Justification: The paper primarily focuses on the experimental exploration of model structures rather than theoretical derivations, hence it does not provide a full set of assumptions or proofs for theoretical results.
    \item[] Guidelines:
    \begin{itemize}
        \item The answer NA means that the paper does not include theoretical results. 
        \item All the theorems, formulas, and proofs in the paper should be numbered and cross-referenced.
        \item All assumptions should be clearly stated or referenced in the statement of any theorems.
        \item The proofs can either appear in the main paper or the supplemental material, but if they appear in the supplemental material, the authors are encouraged to provide a short proof sketch to provide intuition. 
        \item Inversely, any informal proof provided in the core of the paper should be complemented by formal proofs provided in appendix or supplemental material.
        \item Theorems and Lemmas that the proof relies upon should be properly referenced. 
    \end{itemize}

    \item {\bf Experimental Result Reproducibility}
    \item[] Question: Does the paper fully disclose all the information needed to reproduce the main experimental results of the paper to the extent that it affects the main claims and/or conclusions of the paper (regardless of whether the code and data are provided or not)?
    \item[] Answer: \answerYes{} 
    \item[] Justification: The paper fully discloses the necessary information for reproducing the main experimental results, including an anonymized open-source link in the abstract and the use of public datasets. Detailed experimental settings are provided in Sec.~\ref{sec:exp}.
    \item[] Guidelines:
    \begin{itemize}
        \item The answer NA means that the paper does not include experiments.
        \item If the paper includes experiments, a No answer to this question will not be perceived well by the reviewers: Making the paper reproducible is important, regardless of whether the code and data are provided or not.
        \item If the contribution is a dataset and/or model, the authors should describe the steps taken to make their results reproducible or verifiable. 
        \item Depending on the contribution, reproducibility can be accomplished in various ways. For example, if the contribution is a novel architecture, describing the architecture fully might suffice, or if the contribution is a specific model and empirical evaluation, it may be necessary to either make it possible for others to replicate the model with the same dataset, or provide access to the model. In general. releasing code and data is often one good way to accomplish this, but reproducibility can also be provided via detailed instructions for how to replicate the results, access to a hosted model (e.g., in the case of a large language model), releasing of a model checkpoint, or other means that are appropriate to the research performed.
        \item While NeurIPS does not require releasing code, the conference does require all submissions to provide some reasonable avenue for reproducibility, which may depend on the nature of the contribution. For example
        \begin{enumerate}
            \item If the contribution is primarily a new algorithm, the paper should make it clear how to reproduce that algorithm.
            \item If the contribution is primarily a new model architecture, the paper should describe the architecture clearly and fully.
            \item If the contribution is a new model (e.g., a large language model), then there should either be a way to access this model for reproducing the results or a way to reproduce the model (e.g., with an open-source dataset or instructions for how to construct the dataset).
            \item We recognize that reproducibility may be tricky in some cases, in which case authors are welcome to describe the particular way they provide for reproducibility. In the case of closed-source models, it may be that access to the model is limited in some way (e.g., to registered users), but it should be possible for other researchers to have some path to reproducing or verifying the results.
        \end{enumerate}
    \end{itemize}

\item {\bf Open access to data and code}
    \item[] Question: Does the paper provide open access to the data and code, with sufficient instructions to faithfully reproduce the main experimental results, as described in supplemental material?
    \item[] Answer: \answerYes{} 
    \item[] Justification: The paper provides open access to the code through an open-source link included in the abstract, which, along with the description in the supplemental material, offers sufficient instructions to reproduce the main experimental results.
    \item[] Guidelines:
    \begin{itemize}
        \item The answer NA means that paper does not include experiments requiring code.
        \item Please see the NeurIPS code and data submission guidelines (\url{https://nips.cc/public/guides/CodeSubmissionPolicy}) for more details.
        \item While we encourage the release of code and data, we understand that this might not be possible, so “No” is an acceptable answer. Papers cannot be rejected simply for not including code, unless this is central to the contribution (e.g., for a new open-source benchmark).
        \item The instructions should contain the exact command and environment needed to run to reproduce the results. See the NeurIPS code and data submission guidelines (\url{https://nips.cc/public/guides/CodeSubmissionPolicy}) for more details.
        \item The authors should provide instructions on data access and preparation, including how to access the raw data, preprocessed data, intermediate data, and generated data, etc.
        \item The authors should provide scripts to reproduce all experimental results for the new proposed method and baselines. If only a subset of experiments are reproducible, they should state which ones are omitted from the script and why.
        \item At submission time, to preserve anonymity, the authors should release anonymized versions (if applicable).
        \item Providing as much information as possible in supplemental material (appended to the paper) is recommended, but including URLs to data and code is permitted.
    \end{itemize}

\item {\bf Experimental Setting/Details}
    \item[] Question: Does the paper specify all the training and test details (e.g., data splits, hyperparameters, how they were chosen, type of optimizer, etc.) necessary to understand the results?
    \item[] Answer: \answerYes{} 
    \item[] Justification: The paper specifies all the necessary training and test details, including data splits, hyperparameters, their selection process, type of optimizer, etc., both in the experimental section and the appendix, ensuring a clear understanding of the results.
    \item[] Guidelines:
    \begin{itemize}
        \item The answer NA means that the paper does not include experiments.
        \item The experimental setting should be presented in the core of the paper to a level of detail that is necessary to appreciate the results and make sense of them.
        \item The full details can be provided either with the code, in appendix, or as supplemental material.
    \end{itemize}

\item {\bf Experiment Statistical Significance}
    \item[] Question: Does the paper report error bars suitably and correctly defined or other appropriate information about the statistical significance of the experiments?
    \item[] Answer: \answerNo{} 
    \item[] Justification: The paper presents the results from a single run for each experiment, which is consistent with previous works on the same task.
    \item[] Guidelines:
    \begin{itemize}
        \item The answer NA means that the paper does not include experiments.
        \item The authors should answer "Yes" if the results are accompanied by error bars, confidence intervals, or statistical significance tests, at least for the experiments that support the main claims of the paper.
        \item The factors of variability that the error bars are capturing should be clearly stated (for example, train/test split, initialization, random drawing of some parameter, or overall run with given experimental conditions).
        \item The method for calculating the error bars should be explained (closed form formula, call to a library function, bootstrap, etc.)
        \item The assumptions made should be given (e.g., Normally distributed errors).
        \item It should be clear whether the error bar is the standard deviation or the standard error of the mean.
        \item It is OK to report 1-sigma error bars, but one should state it. The authors should preferably report a 2-sigma error bar than state that they have a 96\% CI, if the hypothesis of Normality of errors is not verified.
        \item For asymmetric distributions, the authors should be careful not to show in tables or figures symmetric error bars that would yield results that are out of range (e.g. negative error rates).
        \item If error bars are reported in tables or plots, The authors should explain in the text how they were calculated and reference the corresponding figures or tables in the text.
    \end{itemize}

\item {\bf Experiments Compute Resources}
    \item[] Question: For each experiment, does the paper provide sufficient information on the computer resources (type of compute workers, memory, time of execution) needed to reproduce the experiments?
    \item[] Answer: \answerYes{} 
    \item[] Justification: The paper provides detailed information on the computer resources used for each experiment within Sec.~\ref{sec:exp} and Appendix.
    \item[] Guidelines:
    \begin{itemize}
        \item The answer NA means that the paper does not include experiments.
        \item The paper should indicate the type of compute workers CPU or GPU, internal cluster, or cloud provider, including relevant memory and storage.
        \item The paper should provide the amount of compute required for each of the individual experimental runs as well as estimate the total compute. 
        \item The paper should disclose whether the full research project required more compute than the experiments reported in the paper (e.g., preliminary or failed experiments that didn't make it into the paper). 
    \end{itemize}
    
\item {\bf Code Of Ethics}
    \item[] Question: Does the research conducted in the paper conform, in every respect, with the NeurIPS Code of Ethics \url{https://neurips.cc/public/EthicsGuidelines}?
    \item[] Answer: \answerYes{} 
    \item[] Justification: The research conducted in the paper adheres to the NeurIPS Code of Ethics, ensuring that all aspects of the work, including the methodology, data handling, and reporting, conform to the ethical guidelines provided.
    \item[] Guidelines:
    \begin{itemize}
        \item The answer NA means that the authors have not reviewed the NeurIPS Code of Ethics.
        \item If the authors answer No, they should explain the special circumstances that require a deviation from the Code of Ethics.
        \item The authors should make sure to preserve anonymity (e.g., if there is a special consideration due to laws or regulations in their jurisdiction).
    \end{itemize}

\item {\bf Broader Impacts}
    \item[] Question: Does the paper discuss both potential positive societal impacts and negative societal impacts of the work performed?
    \item[] Answer: \answerYes{} 
    \item[] Justification: The paper includes a discussion on the potential societal impacts of the work in the Appendix.
    \item[] Guidelines:
    \begin{itemize}
        \item The answer NA means that there is no societal impact of the work performed.
        \item If the authors answer NA or No, they should explain why their work has no societal impact or why the paper does not address societal impact.
        \item Examples of negative societal impacts include potential malicious or unintended uses (e.g., disinformation, generating fake profiles, surveillance), fairness considerations (e.g., deployment of technologies that could make decisions that unfairly impact specific groups), privacy considerations, and security considerations.
        \item The conference expects that many papers will be foundational research and not tied to particular applications, let alone deployments. However, if there is a direct path to any negative applications, the authors should point it out. For example, it is legitimate to point out that an improvement in the quality of generative models could be used to generate deepfakes for disinformation. On the other hand, it is not needed to point out that a generic algorithm for optimizing neural networks could enable people to train models that generate Deepfakes faster.
        \item The authors should consider possible harms that could arise when the technology is being used as intended and functioning correctly, harms that could arise when the technology is being used as intended but gives incorrect results, and harms following from (intentional or unintentional) misuse of the technology.
        \item If there are negative societal impacts, the authors could also discuss possible mitigation strategies (e.g., gated release of models, providing defenses in addition to attacks, mechanisms for monitoring misuse, mechanisms to monitor how a system learns from feedback over time, improving the efficiency and accessibility of ML).
    \end{itemize}
    
\item {\bf Safeguards}
    \item[] Question: Does the paper describe safeguards that have been put in place for responsible release of data or models that have a high risk for misuse (e.g., pretrained language models, image generators, or scraped datasets)?
    \item[] Answer: \answerNA{} 
    \item[] Justification: The research exclusively utilizes open-source, public datasets and does not involve high-risk models or scraped data.
    \item[] Guidelines:
    \begin{itemize}
        \item The answer NA means that the paper poses no such risks.
        \item Released models that have a high risk for misuse or dual-use should be released with necessary safeguards to allow for controlled use of the model, for example by requiring that users adhere to usage guidelines or restrictions to access the model or implementing safety filters. 
        \item Datasets that have been scraped from the Internet could pose safety risks. The authors should describe how they avoided releasing unsafe images.
        \item We recognize that providing effective safeguards is challenging, and many papers do not require this, but we encourage authors to take this into account and make a best faith effort.
    \end{itemize}

\item {\bf Licenses for existing assets}
    \item[] Question: Are the creators or original owners of assets (e.g., code, data, models), used in the paper, properly credited and are the license and terms of use explicitly mentioned and properly respected?
    \item[] Answer: \answerYes{} 
    \item[] Justification: The paper credits the creators and original owners of all used assets and explicitly states the licenses and terms of use in Appendix.
    \item[] Guidelines:
    \begin{itemize}
        \item The answer NA means that the paper does not use existing assets.
        \item The authors should cite the original paper that produced the code package or dataset.
        \item The authors should state which version of the asset is used and, if possible, include a URL.
        \item The name of the license (e.g., CC-BY 4.0) should be included for each asset.
        \item For scraped data from a particular source (e.g., website), the copyright and terms of service of that source should be provided.
        \item If assets are released, the license, copyright information, and terms of use in the package should be provided. For popular datasets, \url{paperswithcode.com/datasets} has curated licenses for some datasets. Their licensing guide can help determine the license of a dataset.
        \item For existing datasets that are re-packaged, both the original license and the license of the derived asset (if it has changed) should be provided.
        \item If this information is not available online, the authors are encouraged to reach out to the asset's creators.
    \end{itemize}

\item {\bf New Assets}
    \item[] Question: Are new assets introduced in the paper well documented and is the documentation provided alongside the assets?
    \item[] Answer: \answerNA{} 
    \item[] Justification: The paper does not introduce any new assets.
    \item[] Guidelines:
    \begin{itemize}
        \item The answer NA means that the paper does not release new assets.
        \item Researchers should communicate the details of the dataset/code/model as part of their submissions via structured templates. This includes details about training, license, limitations, etc. 
        \item The paper should discuss whether and how consent was obtained from people whose asset is used.
        \item At submission time, remember to anonymize your assets (if applicable). You can either create an anonymized URL or include an anonymized zip file.
    \end{itemize}

\item {\bf Crowdsourcing and Research with Human Subjects}
    \item[] Question: For crowdsourcing experiments and research with human subjects, does the paper include the full text of instructions given to participants and screenshots, if applicable, as well as details about compensation (if any)? 
    \item[] Answer: \answerNA{} 
    \item[] Justification: The paper does not involve crowdsourcing or human subjects research.
    \item[] Guidelines:
    \begin{itemize}
        \item The answer NA means that the paper does not involve crowdsourcing nor research with human subjects.
        \item Including this information in the supplemental material is fine, but if the main contribution of the paper involves human subjects, then as much detail as possible should be included in the main paper. 
        \item According to the NeurIPS Code of Ethics, workers involved in data collection, curation, or other labor should be paid at least the minimum wage in the country of the data collector. 
    \end{itemize}

\item {\bf Institutional Review Board (IRB) Approvals or Equivalent for Research with Human Subjects}
    \item[] Question: Does the paper describe potential risks incurred by study participants, whether such risks were disclosed to the subjects, and whether Institutional Review Board (IRB) approvals (or an equivalent approval/review based on the requirements of your country or institution) were obtained?
    \item[] Answer: \answerNA{} 
    \item[] Justification: The paper does not involve crowdsourcing or human subjects research.
    \item[] Guidelines:
    \begin{itemize}
        \item The answer NA means that the paper does not involve crowdsourcing nor research with human subjects.
        \item Depending on the country in which research is conducted, IRB approval (or equivalent) may be required for any human subjects research. If you obtained IRB approval, you should clearly state this in the paper. 
        \item We recognize that the procedures for this may vary significantly between institutions and locations, and we expect authors to adhere to the NeurIPS Code of Ethics and the guidelines for their institution. 
        \item For initial submissions, do not include any information that would break anonymity (if applicable), such as the institution conducting the review.
    \end{itemize}

\end{enumerate}

\end{document}